%% file: colm2026_conference.tex
\documentclass{article} 
\usepackage[final]{colm2026_conference}

\usepackage{microtype}
\usepackage{hyperref}
\usepackage{url}
\usepackage{booktabs}
\usepackage{multirow}

\usepackage{lineno}

\definecolor{darkblue}{rgb}{0, 0, 0.5}
\hypersetup{colorlinks=true, citecolor=darkblue, linkcolor=darkblue, urlcolor=darkblue}

\usepackage{graphicx}
\usepackage{subcaption}
\usepackage{tcolorbox}
\usepackage{listings}
\usepackage[T1]{fontenc}
\usepackage{enumitem}
\usepackage{placeins}
\usepackage{makecell}
\usepackage{pifont}
\usepackage[ruled,lined,linesnumbered]{algorithm2e}

\input{math_commands}

\usepackage[capitalize]{cleveref}
\crefformat{section}{\S#2#1#3}
\crefformat{subsection}{\S#2#1#3}
\crefformat{subsubsection}{\S#2#1#3}

\title{Are Latent Reasoning Models Easily Interpretable?}

\author{Connor Dilgren \& Sarah Wiegreffe \\
Department of Computer Science\\
University of Maryland\\
College Park, MD, USA \\
\texttt{\{cdilgren, sarahwie\}@umd.edu}
}

\begin{document}

\ifcolmsubmission
\linenumbers
\fi

\newcommand{\draftonly}[1]{#1}
\renewcommand{\draftonly}[1]{}

\newcommand{\draftcomment}[1]{\draftonly{#1}}
\newcommand{\todo}[1]{\draftcomment{{\textcolor{red}{\small[TODO: #1]}}}}
\newcommand{\sarah}[1]{\draftcomment{{\textcolor{purple}{\small[Sarah: #1]}}}}
\newcommand{\connor}[1]{\draftcomment{{\textcolor{orange}{\small[Connor: #1]}}}}

\maketitle

\begin{abstract}
Latent reasoning models (LRMs) have attracted significant research interest due to their low inference cost (relative to explicit reasoning models) and theoretical ability to explore multiple reasoning paths in parallel. However, these benefits come at the cost of reduced interpretability: LRMs are difficult to monitor because they do not reason in natural language. This paper presents an investigation into LRM interpretability by examining two state-of-the-art LRMs. First, we find that latent reasoning tokens are often unnecessary for LRMs' predictions; on logical reasoning datasets, LRMs can almost always produce the same final answers without using latent reasoning at all. This underutilization of reasoning tokens may partially explain why LRMs do not consistently outperform explicit reasoning methods and raises doubts about the stated role of these tokens in prior work. Second, we demonstrate that when latent reasoning tokens \emph{are} necessary for performance, we can decode gold reasoning traces up to 65-93\% of the time for correctly predicted instances. This suggests LRMs often implement the expected solution rather than an uninterpretable reasoning process. Finally, we present a method to decode a verified natural language reasoning trace from latent tokens without knowing a gold reasoning trace a priori, demonstrating that it is possible to find a verified trace for a majority of correct predictions but only a minority of incorrect predictions. Our findings highlight that current LRMs largely encode interpretable processes, and interpretability itself can be a signal of prediction correctness.


\end{abstract}

\section{Introduction}
Reasoning methods such as chain-of-thought (CoT; \citet{DBLP:journals/corr/abs-2201-11903}) improve the performance of a Language Model (LM) by solving problems in a step-by-step manner.  Theoretical work has demonstrated that reasoning token generation increases the ``effective depth'' of the network by lengthening its longest pathways \citep{feng_towards_2023,li_chain_2023} and helps models solve harder classes of problems \citep{merrill_expressive_2024,nowak_representational_2024,saunshi2025reasoning}. Reasoning token generation has the added benefit of providing users with a form of explanation of models' computational processes in natural language. While the explicit reasoning trace is not always faithful to the model's true reasoning process \citep{wiegreffe-etal-2021-measuring, turpin2023languagemodelsdontsay, chen2025reasoningmodelsdontsay}, it has nonetheless been an important signal for users to calibrate their trust in a model's output \citep{baker2025monitoringreasoningmodelsmisbehavior}.

However, the production of reasoning tokens at inference-time is computationally intensive and many state-of-the art reasoning models (RMs) produce thousands of tokens per query \citep{chen_towards_2025, yeo_demystifying_2025}. An array of recent work has focused on improving RMs' inference-time efficiency \citep{qu_survey_2025, zhu_towards_2025, liu_efficient_2025, sui_stop_2025, feng_efficient_2025, alomrani_reasoning_2025}, with proposed methods ranging from prompting- or decoding-based tricks \citep{wang_wait_2025}, to fine-tuning models to use less reasoning tokens \citep{luo2025o1prunerlengthharmonizingfinetuningo1like}, to dynamically allocating queries based on reasoning necessity \citep{singh2025openaigpt5card}. An approach that has shown promising recent results is that of latent reasoning models (LRMs), which proposes to make reasoning more efficient by forgoing the text decoding process altogether. Methods such as \citet{deng2024explicitcotimplicitcot, hao2025training, deng2025latentreasoningllmsvocabularyspace, cheng2024compressedchainthoughtefficient, geiping2025scaling} train models to autoregressively or recurrently generate additional intermediate latent ``reasoning'' states. Latent reasoning architectures can also be motivated by the intuition that decoding intermediate reasoning hidden states into text is an unneeded bottleneck on information flow \citep{zhu2025surveylatentreasoning}, and theoretical results demonstrating a higher upper-bound on their expressivity \citep{gozeten2025continuouschainthoughtenables, zhu2025reasoning}.

Unfortunately, unlike explicit reasoning models (ERMs), LRMs do not produce human-inspectable reasoning tokens in natural language. This has led to increasing safety concerns about LRMs and calls to preserve explicit reasoning via ``chain-of-thought monitorability'' \citep{korbak2025chainthoughtmonitorabilitynew}. But do we have cause for concern with current LRMs? Prior work \citep{hao2025training, tan2025think} has offered only limited case studies in support of LRM interpretability, with a lack of standardized comparison across architectures or datasets. It is also unclear whether the theoretically demonstrated higher capacity of LRMs has actually been achieved yet with current architectures. We conduct the most comprehensive study to date of latent reasoning interpretability.\footnote{Code available at \url{https://github.com/connordilgren/are-lrms-easily-interpretable}.} We answer three main research questions:

\begin{figure}
    \begin{center}
    \includegraphics[width=1.0\linewidth]{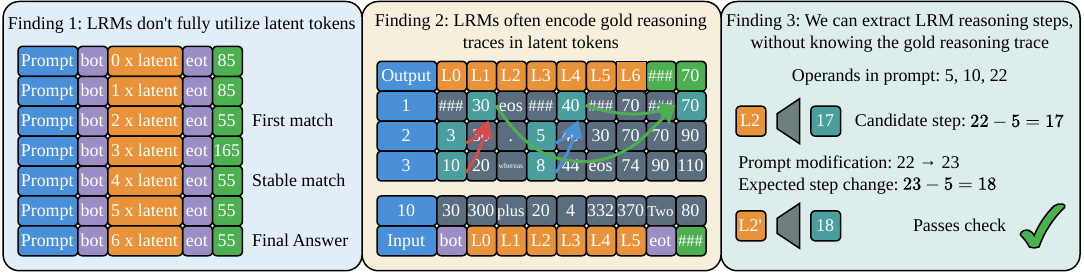}
    \end{center}
    \caption{An overview of our findings. \textbf{Left:} LRMs tend to commit to a final answer before exhausting their budget, indicating that they don't effectively use all available reasoning tokens. \textbf{Middle:} Vocabulary projections of latent tokens often encode gold reasoning traces, suggesting that the model follows an interpretable reasoning trace rather than an opaque one. \textbf{Right:} We can generate candidate steps encoded by a latent token, and verify them by checking whether vocabulary projections changes as expected under modified prompts.}
    \label{fig:figure_1}
\end{figure}

\begin{itemize}[]
    \item[\textbf{RQ1:}] Are latent reasoning tokens necessary for model performance?
    \item[\textbf{RQ2:}] Are gold reasoning traces easily recoverable from latent reasoning tokens?
    \item[\textbf{RQ3:}] Can we extract a reasoning trace that the model is following from latent tokens?
\end{itemize}

We first investigate whether latent reasoning tokens in state-of-the-art LRMs are necessary for performance, a prerequisite for meaningful interpretation. Specifically, we study the width-based LRMs Coconut and CODI (\cref{ssec:lrms}). Somewhat counterintuitively, we find that they are not necessary for some tasks: LRMs' predictions on logical reasoning datasets are almost always the same, regardless of the number of latent tokens available at inference time. We find evidence that the performance gains reported in prior work come from the model's training regimen and not from additional test-time computation.

When models \emph{do} require latent reasoning tokens, we next investigate whether we can find gold reasoning traces encoded in the latent reasoning tokens. We find that it is indeed possible to decode the gold reasoning traces using simple heuristics when models are correct, but less readily when models are incorrect. This suggests that LRMs often follow expected reasoning traces rather than an uninterpretable reasoning process. 

In addition to decoding expected reasoning traces, we present a novel method to decode a verified natural language reasoning trace without knowing a gold reasoning trace a priori. We demonstrate that it is possible to find and verify a reasoning trace for a majority of correct predictions but only a minority of incorrect predictions. Our findings highlight that current LRMs largely encode interpretable processes, and interpretability itself can signal prediction correctness.

\section{Related work}\label{ssec:lrms}\label{ssec:lrm_interp}

\paragraph{Latent reasoning models.} Latent reasoning models reason in continuous hidden states rather than natural language, making their intermediate steps opaque \citep{zhu2025surveylatentreasoning}. We study Coconut \citep{hao2025training} and CODI \citep{shen-etal-2025-codi}, two width-based LRMs that produce intermediate latent reasoning tokens autoregressively and circumvent decoding them into text 
by feeding them directly back into the model as the next token (see \autoref{fig:overview_of_LRMs}).\footnote{We include additional discussion of other types of LRMs in \cref{ssec:rw-extended-lrms}.} We focus on these models because they are increasingly common in the literature, are architecturally similar to ERMs that use chain-of-thought, and have publicly available source code. During training, both the Coconut and CODI models learn to reason from supervision on ground-truth reasoning traces. The Coconut model is instantiated as an ERM; at each stage of the training curriculum, an explicit reasoning step is replaced with a latent reasoning token until no explicit reasoning steps remain. The CODI model trains an ERM alongside the LRM and distills knowledge to the LRM by aligning the hidden states of a key token between the models. We refer the reader to the original papers for more details.

During inference, for both models, a special "beginning of thought" token signals the start of latent reasoning, after which the model processes a predetermined, dataset-specific number of latent tokens. Each latent token is the final-layer hidden state from the previous position, bypassing the standard decoding (and re-embedding) steps of autoregressive generation. An "end of thought" token signals the return to standard decoding for producing the final answer. CODI will additionally pass each final-layer hidden state through a trained two-layer multi-layer perceptron before using it as the next input token during latent reasoning.

From a theoretical angle, recent work \citep{zhu2025reasoning, zou2026the, gozeten2025continuouschainthoughtenables,chen2026the} established higher upper-bound expressivity of width-based LRMs than ERMs, due to the removal of the textual decoding bottleneck. Building off of this, they proposed classes of problems, such as graph reachability or other parallel breadth-first search problems, that LRMs can solve more effectively than ERMs. However, the extent to which \emph{current} LRMs empirically implement these behaviors is unanswered. We find evidence refuting the claim that LRMs exhibit complex search behaviors for certain logical reasoning tasks in \cref{ssec:early-stopping}.

\paragraph{Interpreting latent reasoning models.}
Limited work has been done on interpreting LRMs. Some works proposing LRMs have included interpretability analyses, though largely through case studies. \citet{shen-etal-2025-codi} find preliminary evidence that for correctly-answered math problems, latent tokens can encode intermediate reasoning steps, with step results appearing in the top-$5$ tokens from vocabulary projection and step operands in the top-$10$ attended-to input tokens. \citet{hao2025training} inspect the probabilities assigned to nodes in a graph by latent tokens (after vocabulary projection) and hypothesize that LRMs follow multiple reasoning paths simultaneously. However, it is unclear to what extent these findings hold more generally, or whether they are predictive of models' correctness.

Some concurrent work analyzes LRMs using mechanistic interpretability techniques. \citet{cywinski2025interpret} investigated whether LRMs achieve higher performance than ERMs and non-reasoning models due to latent reasoning or their training regimen. \citet{liang2026latentcotmodelsthinkstepbystep} found that LRMs encode intermediate states in multi-hop tasks with $<3$ hops. We include additional discussion of these works in \cref{ssec:rw-extended-interpret}.

\section{Experimental details}\label{ssec:experimental-details}

\textbf{Datasets.} We perform experiments on three datasets commonly studied in prior work on LRMs: GSM8k-Aug \citep{deng2024implicit}, PrOntoQA \citep[][]{saparov2023languagemodelsgreedyreasoners}, and ProsQA \citep[][]{hao2025training}. See \cref{sec:appendix-dataset-details} for more details, including dataset statistics and examples.

GSM8k-Aug is a dataset of arithmetic problems, each with a 1-8 step gold reasoning trace where every step is an equation that composes operands (e.g., ``3'', ``5'') with operators (e.g., ``+'', ``-'') to produce a result. We add additional valid reasoning traces from the MultiChain GSM8k-Aug dataset \citep{deng2025latentreasoningllmsvocabularyspace}, yielding 1-10 (median 5) gold traces per instance.

PrOntoQA and ProsQA are both logical reasoning datasets that require 6 and 3–6 reasoning steps, respectively. Both tasks require determining whether an entity belongs to a stated category given a set of hierarchical ``is-a'' relationships. ProsQA generally has more distractor paths than PrOntoQA; it was proposed by \citet{hao2025training} to resolve the shortcomings of PrOntoQA for testing search in LRMs.

\begin{table}
\begin{center}
\small
\begin{tabular}{llcccccc}
\toprule
& & \multicolumn{2}{c}{\bf GSM8k-Aug} & \multicolumn{2}{c}{\bf PrOntoQA} & \multicolumn{2}{c}{\bf ProsQA} \\
\cmidrule(lr){3-4} \cmidrule(lr){5-6} \cmidrule(lr){7-8}
\bf Method & \bf Base Model & Acc. (\%) & \# Tok. & Acc. (\%) & \# Tok. & Acc. (\%) & \# Tok. \\
\midrule
No-CoT    & GPT-2 Small & 16.8 (16.5$^\dagger$) & 3.2  & 87.9 (93.8$^\dagger$) & 3.0  & 76.0 (76.7$^\dagger$) & 9.5  \\
CoT       & GPT-2 Small & 41.6 (42.9$^\dagger$) & 31.0 & 99.3 (98.8$^\dagger$) & 92.7 & 74.2 (77.5$^\dagger$) & 51.6 \\
Coconut   & GPT-2 Small & 33.1 (34.1$^\dagger$) & 9.2  & 99.0 (99.8$^\dagger$) & 9.0  & 98.0 (97.0$^\dagger$) & 15.5 \\
CODI      & GPT-2 Small & 42.2$^*$ (43.7$^\ddagger$) & 12.3 & 95.1        & 12.0 & 81.6        & 18.2 \\
\midrule
No-CoT    & Llama-3.2-1B & 30.1 (30.9$^\ddagger$) & 4.2  & 99.8        & 3.0  & 87.8        & 8.6  \\
CoT       & Llama-3.2-1B & 59.4 (61.6$^\ddagger$) & 29.7 & 99.6        & 85.6 & 95.2        & 42.6 \\
Coconut   & Llama-3.2-1B & 35.7 (45.3$^\ddagger$) & 10.2 & 98.8        & 9.0  & 97.6        & 14.7 \\
CODI      & Llama-3.2-1B & 56.0 (55.6$^\ddagger$) & 13.2 & 93.6        & 12.0 & 99.0        & 17.7 \\
\bottomrule
\end{tabular}
\end{center}
\caption{Model performance. Results from \citet{hao2025training}$^\dagger$ and \citet{shen-etal-2025-codi}$^\ddagger$ shown in parentheses where available. See \cref{sec:appendix-table1-discrepancy} for a discussion on the Coconut + Llama-3.2-1B-Instruct performance on GSM8k-Aug compared to the published result.}
\label{tab:performance_results}
\end{table}

\textbf{Models.} Following prior work, we fine-tune GPT-2 Small \citep{radford2019language} and Llama-3.2-1B-Instruct \citep{llama3.2} using the latent training regimens for Coconut and CODI; see \S\ref{ssec:lrms} for details. We additionally fine-tune two baselines: an ERM (i.e., a model that uses chain-of-thought) and a no-CoT model (i.e., a model that answers immediately). We fine-tune each of the four model types separately on each dataset using the training code from \citet{hao2025training,shen-etal-2025-codi}, resulting in twelve models.\footnote{Except for CODI + GPT2-Small on GSM8k-Aug, for which we use the provided checkpoint: \url{https://huggingface.co/zen-E/CODI-gpt2}.} Following \citet{hao2025training} and \citet{shen-etal-2025-codi}, we train and evaluate our Coconut and CODI models using 6 latent reasoning tokens for each dataset. Performance of our replications is in \autoref{tab:performance_results}. Both LRMs outperform the ERM and No-CoT models on ProsQA for both base models.

\section{Are latent reasoning tokens necessary for model performance?}\label{ssec:early-stopping}
We investigate (\cref{ssec:early-stopping-experiment-method}) how effectively LRMs \emph{use} their additional computational power by testing whether their predictions change when latent reasoning is terminated early. If LRMs consistently predict the same answer with fewer latent reasoning tokens, either the task is too easy to test the architecture's benefits, or performance gains stem from the training regimen rather than additional token roll-out. We find evidence for the latter in \cref{ssec:retraining}.

\subsection{Early stopping experiment} \label{ssec:early-stopping-experiment-method}
To determine how many latent reasoning tokens are needed to arrive at a final answer, we prematurely insert the ``end-of-thought'' token to terminate reasoning early and force the model to produce a final answer (see \autoref{fig:figure_1}, left). We then compare predictions using the full $\ell=6$ tokens against reduced counts $\ell \in [0, 1, 2,3,4,5]$ using the metrics:

\begin{enumerate}
    \item \textbf{First match}: the minimum number of reasoning tokens at which the model's answer \emph{matches} its answer given the full set of reasoning tokens. In \autoref{fig:figure_1} (left), the first match occurs at $\ell=2$.
    \item \textbf{Stable match}: the minimum number of reasoning tokens at which the model's answer \emph{remains unchanged} given additional reasoning tokens. In \autoref{fig:figure_1} (left), the stable match occurs at $\ell=4$.
\end{enumerate}

We also run this analysis on the ERMs as a baseline. Since latent reasoning tokens are trained to replace full reasoning steps (\S\ref{ssec:lrms}), we evaluate the ERMs by removing complete steps.

\subsection{Multi-reasoning model experiment}\label{ssec:retraining}
Prior work argues that LRMs achieve high performance on PrOntoQA and ProsQA because they can implement a parallelized breadth-first search at inference time \citep{hao2025training, zhu2025reasoning}. \autoref{tab:performance_results} confirms that LRMs outperform both non-reasoning and explicit reasoning models on ProsQA. However, LRMs also benefit from training on the gold reasoning traces, which the non-reasoning models are not exposed to, and some LRMs (i.e., for GPT-2 Small) make more passes over the training data than the equivalent ERM (see \cref{sec:appendix-model-training} for training parameters). To isolate the effects of additional training data from the architectural modification, we follow the method in \citet{cywinski2025interpret} to train models that are otherwise equivalent to the LRMs in \autoref{tab:performance_results}, but can answer in three modes: no-CoT, explicit reasoning, or latent reasoning. This allows us to directly compare the value of explicit and latent tokens at inference-time when trained on identical data. We train 12 such models across three datasets (GSM8k-Aug, PrOntoQA, ProsQA), two base models (GPT-2 Small, Llama-3.2-1B-Instruct), and two latent reasoning methods (Coconut, CODI). See \cref{sec:appendix-multi-reasoning} for training and inference details for the multi-reasoning models. 

\subsection{Results}

\begin{figure}
    \centering
    \includegraphics{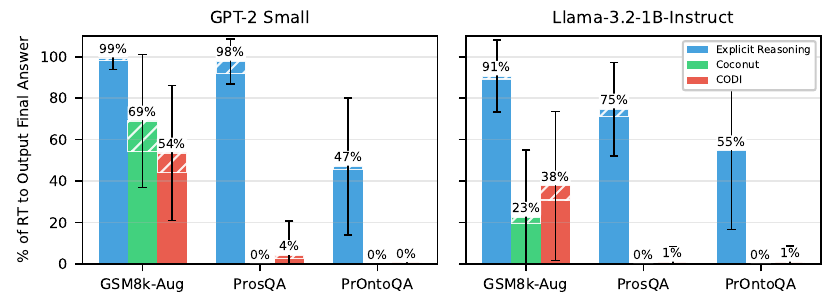}
    \caption{Early stopping results. Solid bars indicate the first match percentage, while hatched bars show the additional reasoning required for a stable match (black lines for one standard deviation), compared to the model's full reasoning trace (RT).}
    \label{fig:early_stopping}
\end{figure}

From the early stopping experiment, we make a surprising finding shown in \autoref{fig:early_stopping}: unlike for GSM8k-Aug, where all models require at least some of their reasoning tokens, LRMs rarely need \emph{any} of their latent reasoning tokens to make stable predictions on PrOntoQA or ProsQA. The ERMs, by comparison, still require 47\% to 98\% of their reasoning tokens. 
This result contradicts the analysis in \cite{hao2025training, zhu2025reasoning}, which argue that the Coconut model uses a parallelized breadth-first search to solve PrOntoQA and ProsQA. See \cref{ssec:comparison-prior-work} for a greater comparison with \cite{hao2025training, zhu2025reasoning}. It is possible that the LRMs perform some form of search either when latent reasoning is not terminated early or as the prompt is being processed, but our result demonstrates that latent tokens are \emph{\textbf{not necessary}} for LRMs to achieve strong performance.\footnote{See \cref{sec:appendix-prontoqa} for a discussion on how models might solve PrOntoQA without learning to search or do first-order logical reasoning.} Future studies should first verify that latent tokens are necessary for a dataset before analyzing how the latent tokens are used by the model.

On GSM8k-Aug, LRMs do use their reasoning tokens, though at lower rates than the explicit model. The underutilization of latent reasoning tokens across all three datasets may partially explain why LRMs do not consistently surpass ERM performance (\autoref{tab:performance_results}). The models' tendency to converge prematurely suggests that they fail to exploit their full computational bandwidth. For this reason, in the subsequent sections, we present results only on the GSM8k-Aug dataset, since establishing latent reasoning tokens' role in performance is a prerequisite to their interpretation. Future work could address underutilization by training models to better use their reasoning budget or by improving efficiency through introducing early stopping mechanisms that terminate reasoning once a stable prediction is reached.

\begin{table}[h]
\begin{center}
\resizebox{\columnwidth}{!}{%
\begin{tabular}{llcccc}
\toprule
\textbf{LRM} & \textbf{Base Model} & \textbf{Reasoning Mode} & \textbf{GSM8k-Aug} & \textbf{PrOntoQA} & \textbf{ProsQA} \\
\midrule
\multirow{4}{*}{Coconut}
 & \multirow{2}{*}{GPT-2 Small} & No-CoT & 0.3 & 0.0 & 0.0 \\
 & & CoT & 19.2 & 0.0 & -6.0 \\
\cmidrule(lr){2-6}
 & \multirow{2}{*}{Llama-3.2-1B-Instruct} & No-CoT & -0.9 & 0.0 & 0.0 \\
 & & CoT & 29.3 & -0.3 & -1.4 \\
\midrule
\multirow{4}{*}{CODI}
 & \multirow{2}{*}{GPT-2 Small} & No-CoT & -5.9 & -0.3 & -0.2 \\
 & & CoT & 3.5 & -0.4 & -7.4 \\
\cmidrule(lr){2-6}
 & \multirow{2}{*}{Llama-3.2-1B-Instruct} & No-CoT & 2.4 & 0.0 & 0.4 \\
 & & CoT & 19.9 & 6.2 & -1.6 \\
\bottomrule
\end{tabular}
}
\end{center}
\caption{Relative performance of non-reasoning and explicit reasoning versus latent reasoning for the multi-reasoning models. Positive values mean latent reasoning underperformed.}
\label{tab:multi-reasoning-performance-rel}
\end{table}

\autoref{tab:multi-reasoning-performance-rel} shows that the apparent advantage of latent reasoning over non-reasoning models on logical reasoning datasets almost disappears when controlling for training data. Coconut has the same performance as no-CoT in PrOntoQA and ProsQA, and CODI is within 0.4 percentage points of no-CoT. Thus, the higher performance of LRMs over non-reasoning models shown in \autoref{tab:performance_results} is likely due to their training regimen and not the additional inference-time compute. Additionally, explicit reasoning continues to outperform latent reasoning in GSM8k-Aug.

\section{Are gold reasoning traces easily recoverable from latent tokens?}\label{sec:backtracking}

When latent reasoning tokens \emph{are} necessary for model performance, can we easily decode gold reasoning traces from them? If so, then LRMs may work as a compressed ERM by solving problems step-by-step in latent space. While prior work has projected latent tokens back to the vocabulary space for interpretation (\S\ref{ssec:lrm_interp}), this has been done either on only a few instances \citep{hao2025training, shen-etal-2025-codi} or in search of intermediate answer quantities rather than the full trace \citep{shen-etal-2025-codi, lu2025latentchainofthoughtdecodingdepthrecurrent}, and only on correct predictions. 

\subsection{Gold reasoning trace backtracking experiment}\label{ssec:backtracking_exp}

\begin{figure}
    \centering
\includegraphics[width=\textwidth]{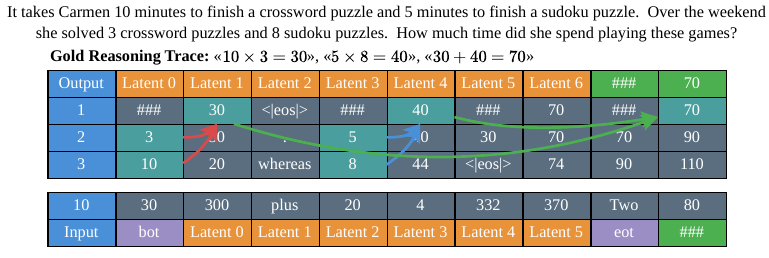}
    \caption{Found gold reasoning trace in Coconut + GPT-2 Small's vocabulary projections, from instance 220 of GSM8k-Aug's test split. The model answered this question correctly.}
    \label{fig:vp_table_sample_220_coconut_success}
\end{figure}

We extract the top-10 tokens from the model's vocabulary\footnote{The top-10 tokens capture at least 90\% of the probability mass over the vocabulary for the median GSM8k-Aug validation instance for Coconut + GPT-2 Small, Coconut + Llama-3.2-1B-Instruct, and CODI + GPT-2 Small. CODI + Llama-3.2-1B-Instruct distributes its probability mass more broadly, such that the top-5000 tokens capture the same probability mass; we use the top-10 tokens for consistency. 
} that each final-layer latent reasoning token projects to using vocabulary projection (i.e., a normalized dot product with the model's unembedding matrix; see \cref{sec:appendix-vocab-projection}).
Making sense of these projections at scale is non-trivial, and they are largely inspected qualitatively in prior work. To rectify this, we devise a backtracking search algorithm to check whether a complete gold reasoning trace is present (\cref{ssec:appendix-backtracking-pseudocode}). Starting from the final step, we verify that the correct answer appears in the top-$k$ tokens at the answer position for incorrect predictions\footnote{\autoref{tab:incorrect-in-topk-all} shows that 46.5\% to 56.0\% of incorrectly predicted instances have the correct answer in the top-10 vocabulary projection at the answer position for all LRMs evaluated.} (this is trivially true for correct predictions). We then recursively check whether each gold reasoning step's operands appear at earlier positions, requiring that operands always precede their results. The trace is considered ``found'' if all steps are located. We run this search both with and without allowing question tokens as operands.\footnote{Consistent with \citet{shen-etal-2025-codi}, we find that LRMs rarely encode operators in vocabulary projections of latent tokens, so we exclude them from the backtracking search.} \autoref{fig:vp_table_sample_220_coconut_success} shows a successfully found reasoning trace. 

As a baseline, we randomly select $n$ reasoning traces from other GSM8k-Aug problems with the same number of steps for each instance. Then, we check whether any of these reasoning traces can also be found using the backtracking search method. If the top-$k$ threshold used in the vocabulary projection is too high, then these random reasoning traces should be found at rates comparable to the gold reasoning traces. We use $n=5$.

\subsection{Results}\label{ssec:backtracking-results}

\begin{figure}
    \centering
    \includegraphics{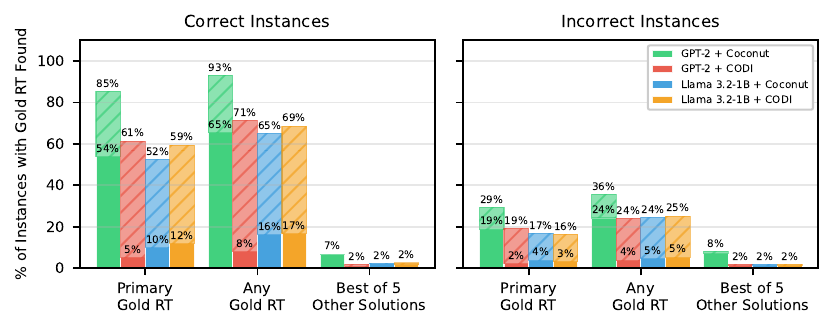}
    \caption{Backtracking results. ``Any Gold RT'' includes additional solutions from the MultiChain GSM8k-Aug dataset \citep{deng2025latentreasoningllmsvocabularyspace}. Darker bars exclude question tokens as operands, and lighter bars show the increase from including them as candidate operands.}
    \label{fig:back_tracking_vp_overall_summary_combined_multi_llm}
\end{figure}

\autoref{fig:back_tracking_vp_overall_summary_combined_multi_llm} shows that the LRMs generally do encode the gold reasoning trace for correctly answered instances. The Coconut + GPT-2 Small model encodes the original gold reasoning trace in 54\% of correctly answered instances. This increases to 65\% when including additional valid reasoning traces from the MultiChain dataset, and then to 93\% when also including numbers from the question as potential operands.

The Coconut + Llama-3.2-1B-Instruct model and both CODI models generally encode intermediate results but not operands into their latent tokens (\autoref{fig:vp_table_sample_036_codi_success}). When question tokens are included as potential operands, at least one gold reasoning trace is found in 65\% to 71\% of correctly answered instances, but this drops to 8\% to 17\% without them.

Somewhat surprisingly, the LRMs sometimes represent the gold reasoning traces even in incorrectly answered problems. The LRMs represent at least one gold reasoning trace 24\% to 36\% of the time when including question tokens as operands. In these cases, an incorrect reasoning trace is encoded more strongly than the gold reasoning trace (\autoref{fig:vp_table_sample_069_coconut_incorrect_GT_found}).

Gold reasoning traces are substantially more represented than random traces from other instances. The best of five random reasoning traces appears only 2\% to 8\% of the time, even when including question tokens as operands. This confirms that the top-10 vocabulary projections are not expressive enough to represent arbitrary reasoning traces.

The results of this experiment provide evidence that LRMs likely solve elementary math problems similarly to ERMs: by calculating intermediate steps and composing them to output a final answer. The main evidence for this is that the gold reasoning traces are consistently present when the model is correct compared to when the model is incorrect, and this is not explained simply by overly expressive vocabulary projections. A plausible explanation is that LRMs learn to compress but still use gold reasoning traces rather than abandoning them for less understandable ways of solving these problems. We note that this evidence is correlational rather than causal; merely finding gold reasoning chains does not prove that they're used by the model.

\section{Can we extract reasoning traces in latent tokens without supervision?}\label{sec:forward_chaining}
The backtracking search method in \cref{ssec:backtracking_exp} checks whether a LRM is encoding a known reasoning trace. But what about interpreting incorrect predictions, where the gold trace may not be present? We propose a second algorithm, forward chaining, to make sense of vocabulary projections when we do not know a gold reasoning trace beforehand.
Our method consists of three steps described below.
See \cref{sec:appendix-forward-chaining} for an example and pseudocode. 
\paragraph{Finding candidate reasoning steps.}For each latent reasoning token, we first find individual reasoning steps that may be encoded. We assume the step result is the top integer token of its vocabulary projection, then find all combinations of operands and arithmetic operators that produce this result, where operands can be results from previous steps, top-$k$ integers from the previous position, or numbers from the prompt.

\paragraph{Verifying candidate reasoning steps.}
To verify that a latent token is encoding a specific candidate step, we create three counterfactual prompts, each with a change to one operand on which that step relies. We next check whether the top integer token of the vocabulary projection corresponding to that step changes to its new expected result; if so, we consider the step ``verified''. 
If not, we try other candidate reasoning steps until none remain. This verification process assumes that the model is robust to minor prompt edits, which can fail if the model restructures its reasoning trace or miscalculates the modified result. To account for this, we vary how many of three verifications must succeed for a step to be verified.


\paragraph{Assembling verified reasoning steps.}
Finally, we assemble found steps into a reasoning trace by starting from the step that produces the final answer and walking backwards, adding steps whose results serve as operands in later steps. A reasoning trace is considered verified if all individual steps are verified. See \autoref{fig:vp_table_sample_290_codi_llama} for an example.

We analyze a 460-instance subset of GSM8k-Aug's test set, filtered for unique, single-token numbers in both the prompt and gold reasoning trace. Unique numbers are required to unambiguously determine which number in the prompt should be modified for verification, and single-token numbers are a limitation of vocabulary projection. See \cref{ssec:appendix-forward-chaining-requirements} for the full set of dataset requirements for forward chaining.

\subsection{Results}

\begin{figure}
    \centering
    \includegraphics{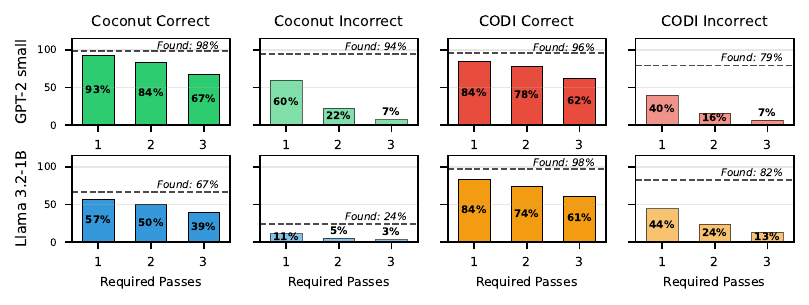}
    \caption{Forward chaining results. 
    }
    \label{fig:forward_chaining_results_2x4}
\end{figure}

As shown in \autoref{fig:forward_chaining_results_2x4}, for Coconut + GPT-2 Small, forward chaining finds and verifies a reasoning trace in 93\% of correctly predicted instances when only requiring $1/3$ verification attempts per step to pass. This drops to 84\% and 67\% when $2/3$ and $3/3$ verification attempts are required to pass, respectively. The LRMs encode verifiable reasoning traces less frequently for incorrectly answered instances. Coconut + GPT-2 Small finds and verifies reasoning traces up to 62 percentage points less for incorrectly predicted instances. This suggests that the LRM does not fully ``show its work'' by skipping one or more steps when the model is incorrect. In doing so, the LRM may be more likely to miscalculate.


For the CODI models, moving from the smaller GPT-2 model (124 million parameters) to the bigger Llama 3.2-1B-Instruct model does not change the percent of found and verified reasoning traces by much ($\leq 8\%$ percentage point difference). In both models, CODI still tends to encode its intermediate results in the top-1 integer token position. But for the Coconut models, moving from GPT-2 to Llama 3.2-1B-Instruct causes up to a 49 percentage point loss in the percent of verified reasoning traces. Coconut + Llama 3.2-1B-Instruct seems to not show its work nearly as much as the Coconut + GPT-2 Small model.



The forward chaining results show that the LRMs studied are moderately interpretable: we can extract and verify a reasoning trace nearly a majority of the time on correct predictions. This is strengthened by the results in \cref{ssec:backtracking-results}. However, this encouraging result may be an artifact of training Coconut and CODI on gold reasoning traces. Standard mechanistic interpretability methods like vocabulary projection may be ineffective on LRMs that have a weaker natural language prior (e.g., models that learn latent reasoning during pretraining). Investigating the interpretability of such models is a promising direction for future work.

As with the gold reasoning traces found via the backtracking experiment in \cref{ssec:backtracking-results}, we do not claim that the reasoning traces found via the forward chaining method are causally used by the model. We believe our method provides a plausible hypothesis on how the model arrived at its final prediction. Finding plausible reasoning traces (with some reason to believe they are implemented, such as through our verification tests) is valuable if they give us some signal into model behavior. For example, prior work has shown that standard chains-of-thought is not always faithful to the model's internal computational process \citep{turpin2023languagemodelsdontsay}, and yet they have still been shown to provide practically useful insights into model behavior, such as reward hacking \citep{baker2025monitoringreasoningmodelsmisbehavior}. In the same way, we hope that the reasoning traces decoded by our forward chaining method will provide important signal into model behavior that can be operationalized in future work to make improvements to model behavior and monitoring.

\section{Conclusion}


This paper investigates LRM interpretability, which is essential for deployment where monitorability is required. Our findings reveal three key insights. First, LRMs do not fully utilize their latent reasoning tokens. On logical reasoning datasets, LRMs often determine their final answer without latent reasoning at all. When controlling for training regimen benefits, no-CoT models match LRM performance on logical reasoning datasets. We encourage future work to investigate on which tasks latent reasoning holds a comparative advantage due to their additional inference-time compute and theoretical capability to follow multiple reasoning traces in parallel. Second, when reasoning tokens are used, gold reasoning traces can be recovered from correct predictions using simple heuristics, suggesting that LRMs implement expected reasoning traces rather than opaque reasoning processes. Finally, we present a method to extract natural language reasoning traces from latent reasoning tokens a majority of the time for correctly-answered instances. Our findings indicate that LRMs are more interpretable than previously assumed, though this may not hold for other classes of LRMs that have a weaker natural language prior.

\bibliography{colm2026_conference}
\bibliographystyle{colm2026_conference}

\appendix

\newpage

\section{Extended related works}\label{sec:appendix-related-works}

\begin{figure}[htbp]
    \centering
    \includegraphics[width=0.9\textwidth]{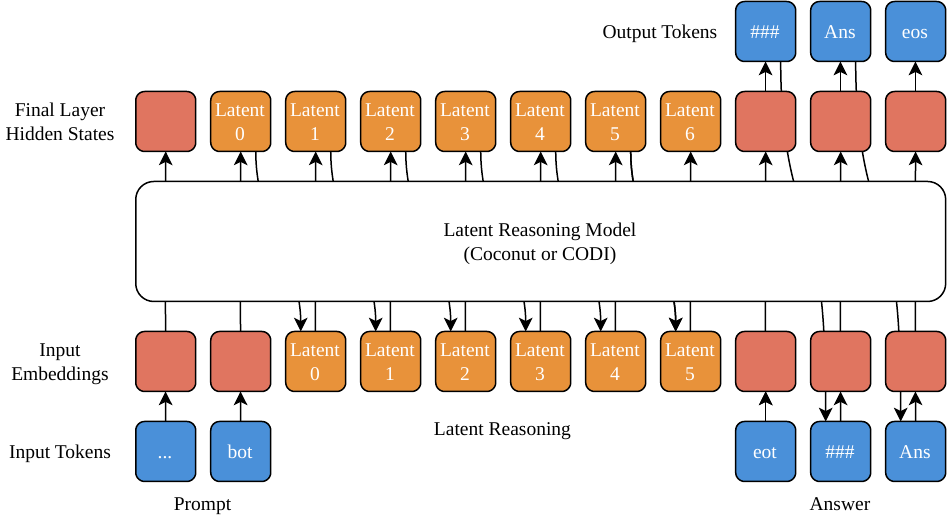}
    \caption{Overview of the latent reasoning models Coconut and CODI. CODI will additionally pass each final-layer hidden state through a trained two-layer multi-layer perceptron before using it as the next input token during latent reasoning.}
    \label{fig:overview_of_LRMs}
\end{figure}

\subsection{Latent reasoning models}\label{ssec:rw-extended-lrms}

Latent reasoning models perform intermediate calculations in a continuous hidden state before answering. This is similar to explicit reasoning models (ERMs) that use chain-of-thought, except the intermediate states are not in natural language and thus aren't readily understandable by humans. This style of architecture has grown in popularity recently; we refer the reader to recent surveys for a more comprehensive overview \citep{zhu2025surveylatentreasoning,chen2025reasoninglanguagecomprehensivesurvey,li2025implicitreasoninglargelanguage}, and related surveys on efficient reasoning \citep{feng_efficient_2025,sui_stop_2025}. 

One of the main paradigms of LRMs, as defined by \citet{zhu2025surveylatentreasoning}, is ``activation-based methods'', which iteratively process representations in a loop. This is exemplified by what we will refer to as ``width-based'' and ``depth-based'' methods.  Depth-based methods \citep[][\emph{inter alia}]{dehghani2018universal,mohtashami2024cotformerchainofthoughtdrivenarchitecture, geiping2025scaling, saunshi2025reasoning,rodkin2025memorizationextendingreasoningdepth,wang2025hierarchicalreasoningmodel,zhu2025scalinglatentreasoninglooped,liu2026thoughtbubblesunsupervisedmethodparallel} generally use layer-wise recurrence to iteratively refine hidden states without expanding the number of ``tokens'' (or ``width''). In this paper, we study width-based LRMs \citep[][\emph{inter alia}]{cheng2024compressedchainthoughtefficient,shen2025hybridcot,deng2026llmlatentreasoningchain,tan2025think}, which alternatively assume a fixed depth defined by the number of layers, but allow models to iteratively generate intermediate latent reasoning tokens autoregressively (illustrated in \autoref{fig:overview_of_LRMs}). 
This is similar to depth-based LRMs, except that by passing the representation at the end of each iteration as the next input token, it allows later positions to attend to it. Specifically, we study the Coconut \citep{hao2025training} and CODI \citep{shen-etal-2025-codi} models. 


\subsection{Interpreting latent reasoning models}\label{ssec:rw-extended-interpret}

Most similar to our work is the recent blogpost by \citet{cywinski2025interpret}, who investigated whether CODI + Llama-3.2-1B-Instruct models achieve higher performance than ERMs and non-reasoning models due to latent reasoning or their training regimen. We follow their method to train a model that can use latent reasoning, CoT, or no-CoT (\cref{ssec:retraining}), in order to expand their analysis to the logical reasoning datasets on which LRMs are found to perform well in prior work \citep{hao2025training, zhu2025reasoning}. \citet{cywinski2025interpret} find that latent reasoning gives a 4.88\% performance uplift over no-CoT on GSM8k-Aug, while we find that latent reasoning decreases performance by 2.4\% (\autoref{tab:multi-reasoning-performance}). In both cases, this is much less than the 24.7\% performance uplift reported in \citet{shen-etal-2025-codi} on separately trained CODI and no-CoT models. Additionally, \citet{cywinski2025interpret} show that vocabulary projection is effective in observing intermediate results stored in latent tokens, and use activation patching to confirm that these intermediate results are needed for model performance. They find a pattern in how CODI solves three step GSM8k-style problems: the third and fifth latent tokens store the first two intermediate results. This regularity enables our forward chaining verification method in \cref{sec:forward_chaining}, which requires a given latent token to compute the same reasoning step across structurally identical problems.

In contrast with other work on LRMs, which tend to focus on LRMs' capacity to follow multiple reasoning traces in parallel, \citet{liang2026latentcotmodelsthinkstepbystep} investigates whether LRMs reason in sequential steps. This work studies the polynomial-iteration dataset \citep{cabannes2024iteration}, which has the (useful for mechanistic interpretability) property of having one valid reasoning trace per instance that uses a small set of possible integers. This work uses vocabulary projection, attention maps, and linear probes to find that the CODI model encodes intermediate states in multi-hop tasks with $\textless 3$ hops, but does not encode the middle intermediates if there are more. They also find that the final step is calculated at the answer token instead of during latent reasoning. 

On the width-based model CoLaR, \citet{tan2025think} show for a specific instance that token embeddings from the gold reasoning trace have high cosine similarity to their LRM's latent tokens. This method is the same as using vocabulary projection (as we do in this work) when the model's embedding matrices are tied.

\citet{peters2025scratchpad} presents evidence that the CODI + GPT-2 model on GSM8k-Aug alternates between localizing operands in one latent token and performing a calculation with those operands in the subsequent latent token. We do not find their results to generalize-- while our vocabulary projection results in \autoref{fig:vp_table_sample_036_codi_success} do show a similar pattern of alternating calculation tokens and non-calculation tokens, we do not observe this pattern for Coconut models (\autoref{fig:vp_table_sample_220_coconut_success}). The CODI + Llama-3.2-1B-Instruct model also seems to have calculation tokens and non-calculation tokens, though not in a clear alternating pattern (\autoref{fig:vp_table_sample_290_codi_llama}).

Though a substantially different architecture from the models we consider, some work has interpreted the hidden states of recurrent-depth LRMs. \citet{geiping2025scaling} perform PCA on the hidden representations of their recurrent-depth architecture, and demonstrate that representations' trajectories during recursion follow distinct geometric patterns. \citet{lu2025latentchainofthoughtdecodingdepthrecurrent} adapt vocabulary projection methods to show evidence against both iterative refinement and structured CoT-like reasoning in the same model, finding that projection to the vocabulary space is less meaningful when depth is increased. 

\subsection{Additional Comparison with Prior Work}\label{ssec:comparison-prior-work}
In this section, we provide more discussion of how our results compare with some of the interpretability experiments in \citet{hao2025training, zhu2025reasoning}. 

Figures 4 and 5 of \citet{hao2025training} support the theorized “latent reasoning as tree search” with a case study that mixes latent and natural language reasoning steps. This case study shows that Coconut can produce mixed reasoning traces of latent tokens followed by natural language steps. This is not surprising, since Coconut models are trained on mixed reasoning traces like this until the final stage of their training curriculum. The measured probability of natural language reasoning steps in their case study does look like a tree search, but they do not show that Coconut needs this reasoning to solve ProsQA. We believe if this study had forced the model to respond with its final answer, rather than to continue a natural language reasoning trace, it would have found the same result as we did, namely that Coconut does not need any reasoning tokens to solve ProsQA.

In Figure 8 of \citet{hao2025training}, they show that, on ProsQA, the first latent token generally places most of its probability mass on the top-1 candidate node, with some probability mass on the second and third candidate nodes. The second latent token concentrates even more probability mass on its top-1 candidate node. Crucially, their analysis does not state what the top-1 candidate node is. Our analysis shows that this top-1 node is actually the model’s final answer! In this interpretation, their result is in agreement with our finding that the Coconut model will converge to its final answer early on in ProsQA rather than exploring multiple parallel paths.

We do share similar conclusions to \citet{hao2025training} in their Figure 9: on GSM8k-Aug, latent tokens do encode reasoning steps, and these steps can be viewed using vocabulary projection. \citet{hao2025training} only shows this for one latent token in one instance; we analyze this phenomenon across $>1000$ instances in \cref{sec:backtracking}.

\citet{zhu2025reasoning} shows a theoretical analysis of how a Coconut model may implement a parallel breadth-first search, where latent token $i$ is the average node embedding of the nodes reachable in $i$ steps. They show empirical evidence for their theory by training a 2-layer, GPT-2 style Coconut model from scratch on a modified version of ProsQA. They show that the inner product between a latent token and node embeddings is higher for reachable nodes than unreachable nodes, which supports their theory. The key difference between their experiment and ours is that their model is trained from scratch. \citet{rizvimartel2026illusionsuperpositionprincipledanalysis} (concurrent with our work) confirm that models trained to use latent reasoning from scratch do implement a parallel breadth-first search on ProsQA, while models fine-tuned to use latent reasoning do not. The extent to which the conclusion of training from scratch holds beyond 2-layer models is unclear.

\section{Dataset details}\label{sec:appendix-dataset-details}

See \autoref{tab:dataset-statistics} for dataset statistics and \autoref{fig:dataset-examples} for examples from each dataset.

For measuring the performance of a model on GSM8k-Aug, we use the original test set of 1,319 instances. For our experiments in \cref{ssec:early-stopping-experiment-method} and \cref{sec:backtracking} and  that require a valid gold reasoning trace, we filter out incomplete instances where the result of the last step in the gold reasoning trace does not match the correct answer, which results in 1,194 test instances. For our experiment in \cref{sec:forward_chaining} which relies on vocabulary projection to extract a reasoning trace, we filter out instances that use multi-token or non-unique numbers, resulting in 460 instances.

\begin{table}[ht]
\begin{center}
\resizebox{\columnwidth}{!}{
\begin{tabular}{lccccc}
\toprule
\multicolumn{1}{c}{\bf Dataset} & \multicolumn{1}{c}{\bf Training} & \multicolumn{1}{c}{\bf Validation} & \multicolumn{1}{c}{\bf Original Test} & \multicolumn{1}{c}{\makecell{\bf Valid Gold\\\bf Reasoning Trace Set}} & \multicolumn{1}{c}{\makecell{\bf VP-Friendly\\\bf Gold Reasoning Trace Set}} \\
\midrule
GSM8k-Aug & 385620 & 500 & 1319 & 1194 & 460 \\
PrOntoQA & 9000 & 200 & 800 & - & - \\
ProsQA & 17886 & 300 & 500 & - & - \\
\bottomrule
\end{tabular}
}
\end{center}
\caption{Dataset statistics. Model performance is calculated on the original test set. The early stopping experiment and backtracking experiment use the valid gold reasoning trace set, which filters out instances where the result of the last reasoning step is not equal to the correct answer. The forward chaining experiment uses the VP (Vocabulary Projection) friendly gold reasoning trace set, which filters for instances that use single-token numbers with unique operands and intermediate results in the gold reasoning trace.}
\label{tab:dataset-statistics}
\end{table}

\begin{figure}[ht]
\centering
\begin{tcolorbox}[
  title=GSM8k-Aug,
  colback=white,
  colframe=black,
  colbacktitle=white,
  coltitle=black,
  fonttitle=\bfseries\small,
  boxrule=0.5pt,
  width=\columnwidth,
  top=4pt, bottom=4pt, left=6pt, right=6pt
]
\small
\textbf{Question:} "Out of 600 employees in a company, 30\% got promoted while 10\% received bonus. How many employees did not get either a promotion or a bonus?"\\
\textbf{Steps:} ["\guillemotleft$600*30/100=180$\guillemotright", "\guillemotleft$600*10/100=60$\guillemotright", "\guillemotleft$180+60=240$\guillemotright", "\guillemotleft$600-240=360$\guillemotright"]\\
\textbf{Answer:} "360"
\end{tcolorbox}

\vspace{6pt}

\begin{tcolorbox}[
  title=PrOntoQA,
  colback=white,
  colframe=black,
  colbacktitle=white,
  coltitle=black,
  fonttitle=\bfseries\small,
  boxrule=0.5pt,
  width=\columnwidth,
  top=4pt, bottom=4pt, left=6pt, right=6pt
]
\small
\textbf{Question:} "Numpuses are not wooden. Vumpuses are lempuses. Rompuses are not dull. Each lorpus is a wumpus. Every gorpus is moderate. Each vumpus is not discordant. Zumpuses are not spicy. Shumpuses are windy. Brimpuses are grimpuses. Each grimpus is a rompus. Brimpuses are zumpuses. Each impus is not opaque. Lorpuses are not mean. Brimpuses are large. Grimpuses are shumpuses. Numpuses are impuses. Shumpuses are numpuses. Lempuses are hot. Numpuses are sterpuses. Shumpuses are gorpuses. Each yumpus is wooden. Every grimpus is orange. Each vumpus is a brimpus. Max is a vumpus. Max is a lorpus. True or false: Max is not wooden."\\
\textbf{Steps:} ["Max is a vumpus. Each vumpus is a brimpus.", "Max is a brimpus. Brimpuses are grimpuses.", "Max is a grimpus. Grimpuses are shumpuses.", "Max is a shumpus. Shumpuses are numpuses.", "Max is a numpus. Numpuses are not wooden.", "Max is not wooden."]\\
\textbf{Answer:} "True"
\end{tcolorbox}

\vspace{6pt}

\begin{tcolorbox}[
  title=ProsQA,
  colback=white,
  colframe=black,
  colbacktitle=white,
  coltitle=black,
  fonttitle=\bfseries\small,
  boxrule=0.5pt,
  width=\columnwidth,
  top=4pt, bottom=4pt, left=6pt, right=6pt
]
\small
Question = "Every kerpus is a yumpus. Every bompus is a boompus. Every vumpus is a felpus. Sally is a vumpus. Every yimpus is a jompus. Every yerpus is a jelpus. Every kerpus is a terpus. Every bompus is a wumpus. Every rempus is a terpus. Every yerpus is a yimpus. Every rempus is a kerpus. Every wumpus is a kerpus. Every impus is a kerpus. Tom is a bompus. Every bompus is a timpus. Sally is a yerpus. Every yumpus is a terpus. Every yumpus is a zhorpus. Every bompus is a impus. Every wumpus is a zhorpus. Every yerpus is a jompus. Every yimpus is a vumpus. Every zumpus is a yumpus. Every zumpus is a rempus. Every zumpus is a storpus. Every timpus is a yumpus. Every impus is a timpus. Every timpus is a rempus. Tom is a impus. Every bompus is a zumpus. Tom is a lempus. Sally is a jompus. Every jelpus is a yimpus. Every rempus is a wumpus. Tom is a rempus. Every yerpus is a vumpus. Every jelpus is a jompus. Every impus is a rempus. Every jelpus is a vumpus. Sally is a storpus. Is Tom a jompus or zhorpus?"\\
Steps = ["Tom is a bompus.", "Every bompus is a wumpus.", "Every wumpus is a zhorpus."]\\
Answer = "Tom is a zhorpus."
\end{tcolorbox}

\caption{Example instances from each dataset.}
\label{fig:dataset-examples}
\end{figure}

\FloatBarrier
\newpage
\section{Model training details}\label{sec:appendix-model-training}

This section details the hyperparameters used to train the models described in \cref{ssec:experimental-details}.

\begin{table}[h]
\begin{center}
\resizebox{\columnwidth}{!}{
\begin{tabular}{lcccccc}
\toprule
& \multicolumn{3}{c}{\bf GPT-2 Small} & \multicolumn{3}{c}{\bf Llama-3.2-1B-Instruct} \\
\cmidrule(lr){2-4} \cmidrule(lr){5-7}
\bf Hyperparameter & GSM8k-Aug & ProsQA & PrOntoQA & GSM8k-Aug & ProsQA & PrOntoQA \\
\midrule
Latent Tokens Per Stage              & 2 & 1 & 1 & 1 & 1 & 1 \\
Stage 0 Epochs                        & 3 & 5 & 5 & 3 & 3 & 3 \\
Epochs Per Stage                      & 3 & 5 & 5 & 1 & 1 & 2 \\
Max Latent Stage                      & 3 & 6 & 6 & 6 & 6 & 6 \\
Total Epochs                          & 50 & 50 & 50 & 10 & 10 & 20 \\
Batch Size                            & 128 & 128 & 128 & 128 & 128 & 128 \\
Learning Rate                         & $1 \times 10^{-4}$ & $1 \times 10^{-4}$ & $1 \times 10^{-4}$ & $5 \times 10^{-5}$ & $5 \times 10^{-5}$ & $5 \times 10^{-5}$ \\
Weight Decay                          & 0.01 & 0.01 & 0.01 & 0.1 & 0.1 & 0.1 \\
BF16 Precision                        & \ding{55} & \ding{55} & \ding{55} & \ding{51} & \ding{51} & \ding{51} \\
Reset Optimizer Between Stages        & \ding{51} & \ding{51} & \ding{51} & \ding{51} & \ding{51} & \ding{51} \\
\bottomrule
\end{tabular}
}
\end{center}
\caption{Training hyperparameters for \textbf{Coconut} models. For the Coconut + GPT-2 Small model on GSM8k-Aug, the stage 0 training is replaced with checkpoint 6 of the CoT model.}\label{tab:coconut-hyperparameters}
\end{table}

\begin{table}[h]
\begin{center}
\resizebox{\columnwidth}{!}{
\begin{tabular}{lcccccc}
\toprule
& \multicolumn{3}{c}{\bf GPT-2 Small} & \multicolumn{3}{c}{\bf Llama-3.2-1B-Instruct} \\
\cmidrule(lr){2-4} \cmidrule(lr){5-7}
\bf Hyperparameter & GSM8k-Aug & ProsQA & PrOntoQA & GSM8k-Aug & ProsQA & PrOntoQA \\
\midrule
Latent Loss Weight ($\alpha$)        & 1 & 1 & 1 & 1 & 1 & 1 \\
CoT Loss Weight ($\beta$)            & 1 & 1 & 1 & 1 & 1 & 1 \\
Distillation Loss Weight ($\delta$)  & 1 & 1 & 1 & 20 & 20 & 20 \\
Num Latent Tokens                     & 6 & 6 & 6 & 6 & 6 & 6 \\
Total Epochs                          & 40 & 40 & 40 & 10 & 10 & 10 \\
Batch Size                            & 128 & 128 & 128 & 128 & 128 & 128 \\
Learning Rate                         & $3 \times 10^{-3}$ & $3 \times 10^{-3}$ & $3 \times 10^{-3}$ & $8 \times 10^{-4}$ & $8 \times 10^{-4}$ & $8 \times 10^{-4}$ \\
Weight Decay                          & 0.1 & 0.1 & 0.1 & 0.1 & 0.1 & 0.1 \\
BF16 Precision                        & \ding{51} & \ding{51} & \ding{51} & \ding{51} & \ding{51} & \ding{51} \\
Projection Dim                        & 768 & 768 & 768 & 2048 & 2048 & 2048 \\
LoRA Rank                             & 128 & 128 & 128 & 128 & 128 & 128 \\
LoRA Alpha                            & 32 & 32 & 32 & 32 & 32 & 32 \\
\bottomrule
\end{tabular}
}
\end{center}
\caption{Training hyperparameters for \textbf{CODI} models. For the CODI + GPT-2 Small model on GSM8k-Aug, we used a checkpoint released by the \citet{shen-etal-2025-codi} authors.}\label{tab:codi-hyperparameters}
\end{table}

\begin{table}[h]
\begin{center}
\resizebox{\columnwidth}{!}{
\begin{tabular}{lcccccc}
\toprule
& \multicolumn{3}{c}{\bf GPT-2 Small} & \multicolumn{3}{c}{\bf Llama-3.2-1B-Instruct} \\
\cmidrule(lr){2-4} \cmidrule(lr){5-7}
\bf Hyperparameter & GSM8k-Aug & ProsQA & PrOntoQA & GSM8k-Aug & ProsQA & PrOntoQA \\
\midrule
Total Epochs                          & 25 & 25 & 25 & 10 & 10 & 10 \\
Batch Size                            & 128 & 128 & 128 & 128 & 128 & 128 \\
Learning Rate                         & $1 \times 10^{-4}$ & $1 \times 10^{-4}$ & $1 \times 10^{-4}$ & $5 \times 10^{-5}$ & $5 \times 10^{-5}$ & $5 \times 10^{-5}$ \\
Weight Decay                          & 0.01 & 0.01 & 0.01 & 0.1 & 0.1 & 0.1 \\
BF16 Precision                        & \ding{55} & \ding{55} & \ding{55} & \ding{51} & \ding{51} & \ding{51} \\
\bottomrule
\end{tabular}
}
\end{center}
\caption{Training hyperparameters for \textbf{CoT} models.}\label{tab:cot-hyperparameters}
\end{table}

\begin{table}[h]
\begin{center}
\resizebox{\columnwidth}{!}{
\begin{tabular}{lcccccc}
\toprule
& \multicolumn{3}{c}{\bf GPT-2 Small} & \multicolumn{3}{c}{\bf Llama-3.2-1B-Instruct} \\
\cmidrule(lr){2-4} \cmidrule(lr){5-7}
\bf Hyperparameter & GSM8k-Aug & ProsQA & PrOntoQA & GSM8k-Aug & ProsQA & PrOntoQA \\
\midrule
Total Epochs                          & 25 & 25 & 25 & 10 & 10 & 20 \\
Batch Size                            & 128 & 128 & 128 & 128 & 128 & 128 \\
Learning Rate                         & $1 \times 10^{-4}$ & $1 \times 10^{-4}$ & $1 \times 10^{-4}$ & $5 \times 10^{-5}$ & $5 \times 10^{-5}$ & $5 \times 10^{-5}$ \\
Weight Decay                          & 0.01 & 0.01 & 0.01 & 0.1 & 0.1 & 0.1 \\
BF16 Precision                        & \ding{55} & \ding{55} & \ding{55} & \ding{51} & \ding{51} & \ding{51} \\
\bottomrule
\end{tabular}
}
\end{center}
\caption{Training hyperparameters for \textbf{No-CoT} models.}\label{tab:nocot-hyperparameters}
\end{table}

\FloatBarrier
\newpage
\section{Multi-reasoning model training details}\label{sec:appendix-multi-reasoning}

We train the multi-reasoning models that use Coconut for latent reasoning using the following loss function, which is the same as the original Coconut loss function but with the $\beta \mathcal{L}_{\text{explicit reasoning}}$ and $\gamma \mathcal{L}_{\text{direct answer}}$ terms added:

\begin{equation}
\mathcal{L}_{\text{multi-reasoning, Coconut}} = \alpha \mathcal{L}_{\text{latent reasoning}} + \beta \mathcal{L}_{\text{explicit reasoning}} + \gamma \mathcal{L}_{\text{direct answer}}
\label{eq:total_loss_coconut_multi-reasoning}
\end{equation}

We train the multi-reasoning models that use CODI for latent reasoning using the following loss function, which is the same as the original CODI loss function but with the $\gamma \mathcal{L}_{\text{direct answer}}$ term added:

\begin{equation}
\mathcal{L}_{\text{multi-reasoning, CODI}} = \alpha \mathcal{L}_{\text{latent reasoning}} + \beta \mathcal{L}_{\text{explicit reasoning}} + \gamma \mathcal{L}_{\text{direct answer}} + \delta \mathcal{L}_{\text{KD}}
\label{eq:total_loss_codi_multi-reasoning}
\end{equation}

We select the model checkpoint with the highest harmonic mean across the three reasoning modes on the validation set.

\autoref{tab:multi-reasoning-hyperparameters-coconut} and \autoref{tab:multi-reasoning-hyperparameters-codi} show the hyperparameters used to train the multi-reasoning models. In most cases, they are the same hyperparameters as the original single reasoning mode models. At inference time, we control the reasoning mode using the control tokens in \autoref{tab:control-tokens}. The test-set accuracy of the multi-reasoning models for each reasoning mode is in \autoref{tab:multi-reasoning-performance}.

\begin{table}[h]
\begin{center}
\resizebox{\columnwidth}{!}{
\begin{tabular}{lcccccc}
\toprule
& \multicolumn{3}{c}{\bf GPT-2 Small} & \multicolumn{3}{c}{\bf Llama-3.2-1B-Instruct} \\
\cmidrule(lr){2-4} \cmidrule(lr){5-7}
\bf Hyperparameter & GSM8k-Aug & ProsQA & PrOntoQA & GSM8k-Aug & ProsQA & PrOntoQA \\
\midrule
Latent Loss Weight ($\alpha$)       & 1 & 1 & 1 & 1 & 1 & 1 \\
CoT Loss Weight ($\beta$)           & 1 & 1 & 1 & 1 & 1 & 1 \\
No-CoT Loss Weight ($\gamma$)       & 1 & 1 & 1 & 1 & 1 & 1 \\
Latent Tokens Per Stage              & 2 & 1 & 1 & 1 & 1 & 1 \\
Stage 0 Epochs                        & 3 & 5 & 5 & 3 & 3 & 3 \\
Epochs Per Stage                      & 3 & 5 & 5 & 1 & 1 & 2 \\
Max Latent Stage                      & 3 & 6 & 6 & 6 & 6 & 6 \\
Total Epochs                          & 50 & 50 & 50 & 10 & 10 & 20 \\
Batch Size                            & 128 & 128 & 128 & 128 & 128 & 128 \\
Learning Rate                         & $1 \times 10^{-4}$ & $1 \times 10^{-4}$ & $1 \times 10^{-4}$ & $5 \times 10^{-5}$ & $5 \times 10^{-5}$ & $5 \times 10^{-5}$ \\
Weight Decay                          & 0.01 & 0.01 & 0.01 & 0.1 & 0.1 & 0.1 \\
BF16 Precision                        & \ding{55} & \ding{55} & \ding{55} & \ding{51} & \ding{51} & \ding{51} \\
Reset Optimizer Between Stages        & \ding{51} & \ding{51} & \ding{51} & \ding{51} & \ding{51} & \ding{51} \\
\bottomrule
\end{tabular}
}
\end{center}
\caption{Training hyperparameters for multi-reasoning mode models which use \textbf{Coconut} for latent reasoning.}\label{tab:multi-reasoning-hyperparameters-coconut}
\end{table}

\begin{table}[h]
\begin{center}
\resizebox{\columnwidth}{!}{
\begin{tabular}{lcccccc}
\toprule
& \multicolumn{3}{c}{\bf GPT-2 Small} & \multicolumn{3}{c}{\bf Llama-3.2-1B-Instruct} \\
\cmidrule(lr){2-4} \cmidrule(lr){5-7}
\bf Hyperparameter & GSM8k-Aug & ProsQA & PrOntoQA & GSM8k-Aug & ProsQA & PrOntoQA \\
\midrule
Latent Loss Weight ($\alpha$)       & 1 & 1 & 1 & 1 & 1 & 1 \\
CoT Loss Weight ($\beta$)           & 1 & 1 & 1 & 1 & 1 & 1 \\
No-CoT Loss Weight ($\gamma$)       & 1 & 1 & 1 & 1 & 1 & 1 \\
Distillation Loss Weight ($\delta$)      & 1 & 1 & 1 & 20 & 20 & 20 \\
Num Latent Tokens                     & 6 & 6 & 6 & 6 & 6 & 6 \\
Total Epochs                          & 40 & 40 & 40 & 10 & 10 & 20 \\
Batch Size                            & 128 & 128 & 128 & 128 & 128 & 128 \\
Learning Rate                         & $3 \times 10^{-3}$ & $3 \times 10^{-3}$ & $3 \times 10^{-3}$ & $8 \times 10^{-4}$ & $8 \times 10^{-4}$ & $8 \times 10^{-4}$ \\
Weight Decay                          & 0.1 & 0.1 & 0.1 & 0.1 & 0.1 & 0.1 \\
BF16 Precision                        & \ding{51} & \ding{51} & \ding{51} & \ding{51} & \ding{51} & \ding{51} \\
Projection Dim                        & 768 & 768 & 768 & 2048 & 2048 & 2048 \\
LoRA Rank                             & 128 & 128 & 128 & 128 & 128 & 128 \\
LoRA Alpha                            & 32 & 32 & 32 & 32 & 32 & 32 \\
\bottomrule
\end{tabular}
}
\end{center}
\caption{Training hyperparameters for multi-reasoning mode models which use \textbf{CODI} for latent reasoning.}\label{tab:multi-reasoning-hyperparameters-codi}
\end{table}

\begin{table}[h]
\begin{center}
\begin{tabular}{lll}
\toprule
\textbf{Base Model} & \textbf{Mode} & \textbf{Control Tokens} \\
\midrule
\multirow{3}{*}{GPT-2 Small} & No-CoT      & [prompt] + [eocot] $\rightarrow$ answer \\
                        & CoT  & [prompt] + [bocot] $\rightarrow$ cot + eocot $\rightarrow$ answer \\
                        & Latent      & [prompt] + [bocot] $\rightarrow$ latent $\rightarrow$ [eocot] $\rightarrow$ answer \\
\midrule
\multirow{3}{*}{Llama-3.2-1B}  & No-CoT      & [prompt] + [eot] + [eocot] $\rightarrow$ answer \\
                        & CoT  & [prompt] + [eot] + [bocot] $\rightarrow$ cot + eocot $\rightarrow$ answer \\
                        & Latent      & [prompt] + [eot] + [bocot] $\rightarrow$ latent $\rightarrow$ [eocot] $\rightarrow$ answer \\
\bottomrule
\end{tabular}
\end{center}
\caption{Control tokens used to determine which reasoning mode the model uses. Tokens in square brackets are model inputs; they are not generated by the model. In CoT mode, the ``eocot'' is output by the model, while in latent mode, this token is an input. Llama-3.2-1B-Instruct uses the additional ``end-of-turn'' token after the prompt.}
\label{tab:control-tokens}
\end{table}

\begin{table}[h]
\begin{center}
\resizebox{\columnwidth}{!}{%
\begin{tabular}{llcccc}
\toprule
\textbf{LRM} & \textbf{Base Model} & \textbf{Reasoning Mode} & \textbf{GSM8k-Aug} & \textbf{PrOntoQA} & \textbf{ProsQA} \\
\midrule
\multirow{6}{*}{Coconut}
 & \multirow{3}{*}{GPT-2 Small} & No-CoT & 22.4 & 100.0 & 94.2 \\
 & & CoT & 41.3 & 100.0 & 88.2 \\
 & & Coconut & 22.1 & 100.0 & 94.2 \\
\cmidrule(lr){2-6}
 & \multirow{3}{*}{Llama-3.2-1B-Instruct} & No-CoT & 29.2 & 94.9 & 99.8 \\
 & & CoT & 59.4 & 94.6 & 98.4 \\
 & & Coconut & 30.1 & 94.9 & 99.8 \\
\midrule
\multirow{6}{*}{CODI}
 & \multirow{3}{*}{GPT-2 Small} & No-CoT & 27.8 & 94.5 & 79.8 \\
 & & CoT & 37.2 & 94.4 & 72.6 \\
 & & CODI & 33.7 & 94.8 & 80.0 \\
\cmidrule(lr){2-6}
 & \multirow{3}{*}{Llama-3.2-1B-Instruct} & No-CoT & 36.1 (36.7) & 93.4 & 96.8 \\
 & & CoT & 53.6 (42.1) & 99.6 & 94.8 \\
 & & CODI & 33.7 (41.6) & 93.4 & 96.4 \\
\bottomrule
\end{tabular}
}
\end{center}
\caption{Multi-reasoning model accuracy for each reasoning mode. This is the same data as in \autoref{tab:multi-reasoning-performance-rel}, but with absolute performance instead of relative. Results from \citet{cywinski2025interpret} shown in parentheses where available. }\label{tab:multi-reasoning-performance}
\end{table}

\FloatBarrier
\newpage
\section{Early stopping experiment results}\label{sec:appendix-early-stopping}

\begin{table}[h]
\begin{center}
\small
\begin{tabular}{llcccccc}
\toprule
& & \multicolumn{3}{c}{\bf First Match (\%)} & \multicolumn{3}{c}{\bf Stable Match (\%)} \\
\cmidrule(lr){3-5} \cmidrule(lr){6-8}
\bf Base Model & \bf Dataset & \bf Explicit & \bf Coconut & \bf CODI & \bf Explicit & \bf Coconut & \bf CODI \\
\midrule
\multirow{3}{*}{GPT-2 Small}
 & GSM8k-Aug    & 98.1\scriptsize{$\pm$9.8}  & 54.4\scriptsize{$\pm$36.2} & 44.2\scriptsize{$\pm$33.6} & 99.3\scriptsize{$\pm$5.4}  & 69.0\scriptsize{$\pm$32.1} & 53.6\scriptsize{$\pm$32.5} \\
 & ProsQA   & 92.0\scriptsize{$\pm$21.7} & 0.0\scriptsize{$\pm$0.0}   & 2.5\scriptsize{$\pm$12.6}  & 97.7\scriptsize{$\pm$10.9} & 0.0\scriptsize{$\pm$0.0}   & 4.0\scriptsize{$\pm$16.6}  \\
 & PrOntoQA & 45.5\scriptsize{$\pm$32.8} & 0.1\scriptsize{$\pm$2.9}   & 0.3\scriptsize{$\pm$4.4}   & 47.1\scriptsize{$\pm$33.0} & 0.1\scriptsize{$\pm$2.9}   & 0.5\scriptsize{$\pm$5.8}   \\
\midrule
\multirow{3}{*}{Llama-3.2-1B}
 & GSM8k-Aug    & 89.3\scriptsize{$\pm$19.1} & 19.5\scriptsize{$\pm$30.0} & 30.9\scriptsize{$\pm$33.9} & 90.6\scriptsize{$\pm$17.4} & 22.5\scriptsize{$\pm$32.5} & 37.7\scriptsize{$\pm$35.9} \\
 & ProsQA   & 71.1\scriptsize{$\pm$22.6} & 0.0\scriptsize{$\pm$0.7}   & 0.1\scriptsize{$\pm$2.2}   & 74.7\scriptsize{$\pm$22.5} & 0.0\scriptsize{$\pm$0.7}   & 0.7\scriptsize{$\pm$7.4}   \\
 & PrOntoQA & 54.6\scriptsize{$\pm$38.0} & 0.0\scriptsize{$\pm$0.6}   & 0.3\scriptsize{$\pm$4.8}   & 54.7\scriptsize{$\pm$38.0} & 0.0\scriptsize{$\pm$0.6}   & 0.7\scriptsize{$\pm$8.0}   \\
\bottomrule
\end{tabular}
\end{center}
\caption{Early stopping experiment results. This is the same data as \autoref{fig:early_stopping}.}\label{early-stopping-tabletable}
\end{table}

\FloatBarrier
\section{Vocabulary projection details}\label{sec:appendix-vocab-projection}
We use the popular vocabulary projection technique (or ``logit lens''; \citet{nostalgebraist2020logitlens, geva_transformer_2021}) to map latent tokens back to the model's vocabulary space. This is done by multiplying the residual stream after the final layer (and final LayerNorm) with the model's unembedding matrix to obtain an (unnormalized) distribution over the vocabulary. We repeat this at each latent token position, obtaining the top-$k$ natural language tokens (i.e., rows of the unembedding matrix) with the highest dot product against each latent token; this is equivalent to how a natural language token would be decoded should the model have been operating as an ERM.

Vocabulary projection, used in \cref{sec:backtracking} and \cref{sec:forward_chaining}, only reveals single-token concepts; it omits multi-token concepts and latent space directions not well-aligned with vocabulary space. We encourage future work to develop core mechanistic interpretability tools that can address these limitations, which would make LRMs more interpretable.

To account for vocabulary projection's single-token limitation, in \cref{sec:backtracking}, we assume that the first non-zero integer token of a multi-token number represents the full number. E.g., we assume ``0.5'' is represented by ``5''. 

\FloatBarrier
\section{Coconut + Llama-3.2-1B-Instruct performance}\label{sec:appendix-table1-discrepancy}
The published performance results in \autoref{tab:performance_results} are close to our models' performance, except for the Coconut + Llama-3.2-1B-Instruct model trained on GSM8k-Aug, where our model performs 9.6 percentage points worse. Even though this is a Coconut model, the published result comes from \citet{shen-etal-2025-codi}, since \citet{hao2025training} did not train a Coconut + Llama-3.2-1B-Instruct model. It's likely that \citet{shen-etal-2025-codi} trained their model with a different set of hyperparameters. We believe our performance result of 35.7\% is trustworthy since this is in the ballpark of the 31.7\% performance result reported by \citet{hao2025training} for a Coconut + Llama 3.2-3B model on GSM8k-Aug.

\FloatBarrier
\section{Additional Experiments on Latent Reasoning Usefulness}\label{sec:appendix-additional-usefulness-experiments}

The early stopping experiment in \cref{ssec:early-stopping-experiment-method} showed that latent reasoning is not necessary for a model to arrive at its final answer on ProsQA and PrOntoQA. But it's possible that latent reasoning is still useful in someway, such as by increasing the probability of the correct answer, even though the model's top-1 prediction on the final answer is not changing. We run the following additional experiments to provide evidence that the latent tokens have no useful information for ProsQA and PrOntoQA, though they do for GSM8k-Aug.

We replace the latent tokens during runtime with a mean latent token calculated at each position using the validation set for GSM8k-Aug, PrOntoQA, and ProsQA. If latent tokens contain instance-specific information, then replacing them with these mean latent tokens should remove that information and decrease model performance. \autoref{tab:mean-ablated-rel-accuracy} shows that, for PrOntoQA and ProsQA, the test set accuracy hardly changes, by only -0.1 to +0.2 percentage points. On GSM8k-Aug, model performance dropped 5.8 to 31.5 percentage points. This is additional evidence that the latent reasoning tokens do not contain useful information for making predictions on PrOntoQA and ProsQA.

\begin{table}
\begin{center}
\small
\begin{tabular}{llccc}
\toprule
\bf Method & \bf Base Model & \bf GSM8k-Aug & \bf PrOntoQA & \bf ProsQA \\
 & & $\Delta$ Acc. (\%) & $\Delta$ Acc. (\%) & $\Delta$ Acc. (\%) \\
\midrule
Coconut & GPT-2 Small  & $-26.0$ & $0.0$   & $0.0$ \\
CODI    & GPT-2 Small  & $-31.5$ & $0.2$   & $0.2$ \\
\midrule
Coconut & Llama-3.2-1B & $-5.8$  & $0.0$   & $0.0$ \\
CODI    & Llama-3.2-1B & $-25.0$ & $-0.1$  & $0.0$ \\
\bottomrule
\end{tabular}
\end{center}
\caption{Change in accuracy when using mean latent tokens. Negative values indicate performance is lost when using mean latent tokens.}
\label{tab:mean-ablated-rel-accuracy}
\end{table}

\begin{figure}[htbp]
    \centering
    \includegraphics[width=1.0\textwidth]{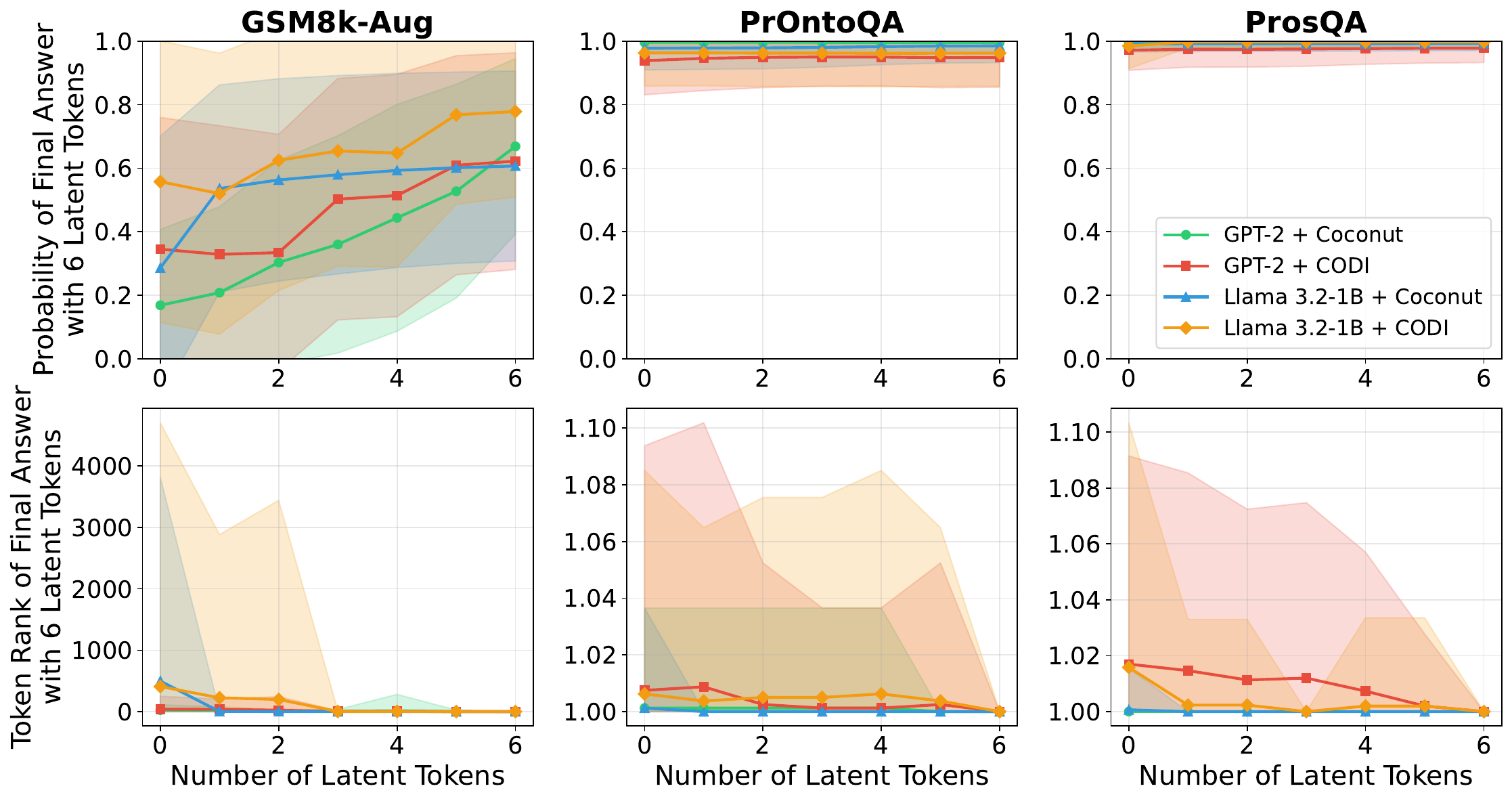}
    \caption{Probability and token rank of the model's final answer during latent reasoning.}
    \label{fig:model_confidence}
\end{figure}

Additionally, we measure the probability and token rank of the model’s final answer (on a standard run with 6 latent tokens) given $k \in \{0, \dots, 6\}$ latent tokens. If the latent tokens do encode something useful, we’d expect the final answer’s probability and token rank to move towards 1. \autoref{fig:model_confidence} shows that on PrOntoQA and ProsQA, the probability and token rank of the model’s final answer remains near 1 on average regardless of how many latent tokens were used. On GSM8k-Aug, the probability and token rank of the model’s final answer do improve as more latent tokens are used. This is additional evidence that the latent tokens provide useful information for GSM8k-Aug, but not PrOntoQA or ProsQA.

\begin{figure}[htbp]
    \centering
    \includegraphics[width=1.0\textwidth]{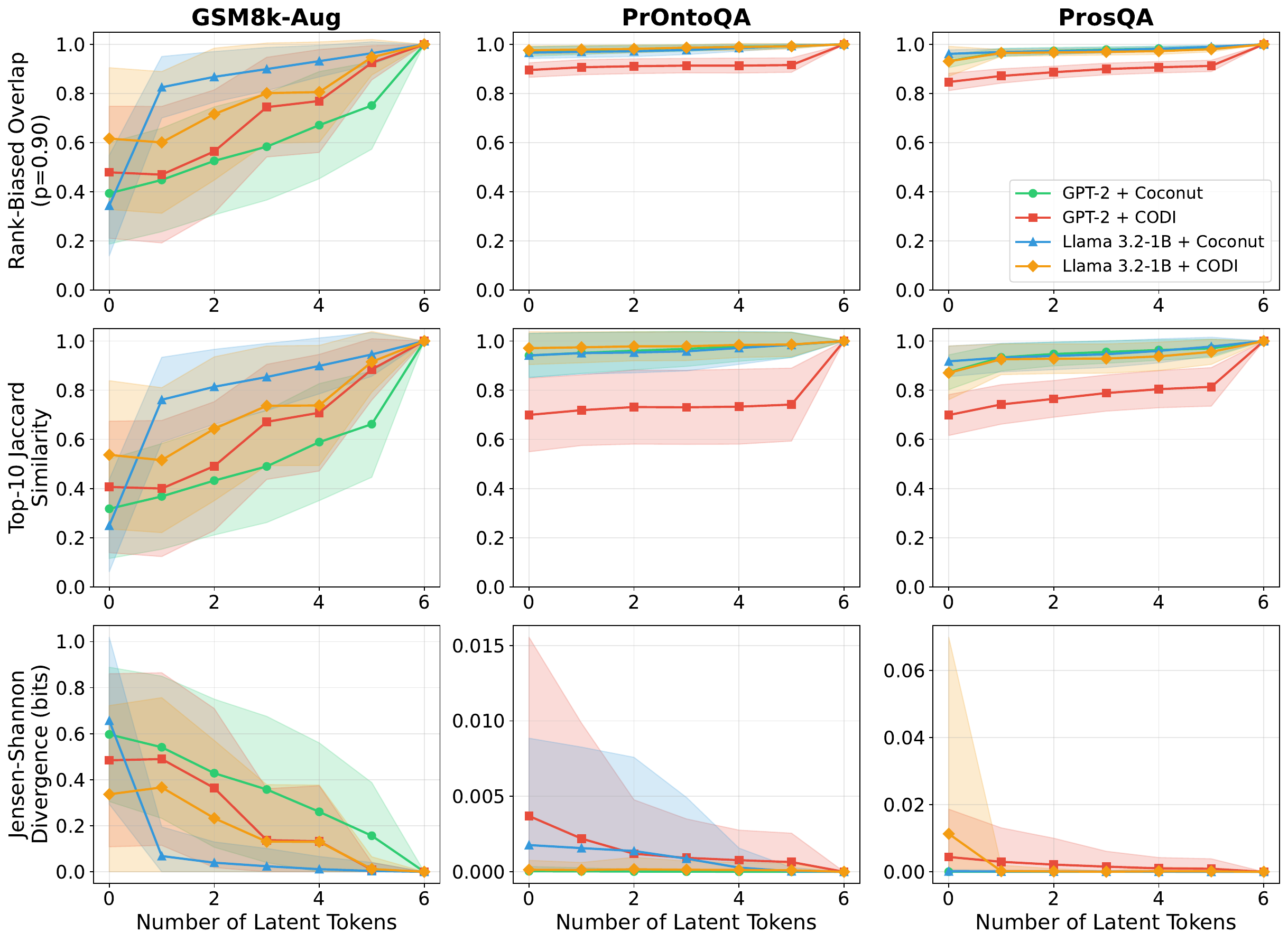}
    \caption{Rank-Biased Overlap (p=0.90), Top-k Jaccard similarity (k=10), and Jensen-Shannon Divergence between the model's forced early answer distribution and its final answer distribution.}
    \label{fig:distributional_similarity}
\end{figure}

\begin{figure*}[htbp]
    \centering
    \includegraphics[width=\textwidth]{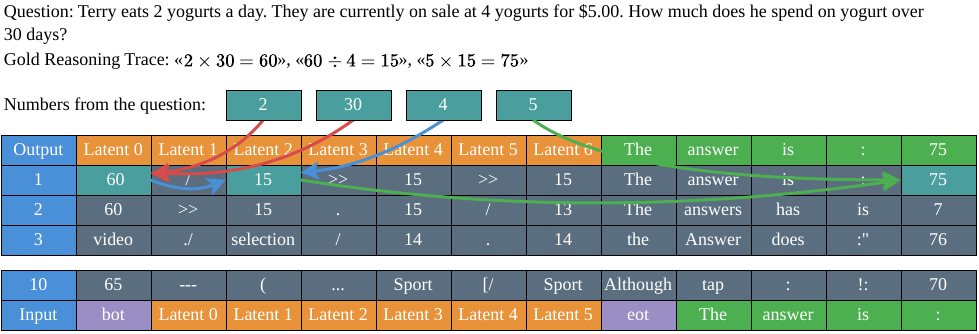}
    \caption{Found gold reasoning trace in CODI + GPT-2 Small's vocabulary projections, from instance 36 of GSM8k-Aug's test split. The CODI model does not encode numbers from the question in the latent tokens, at least not in a way that is detectable using vocabulary projection. The model answered this question correctly.}
    \label{fig:vp_table_sample_036_codi_success}
\end{figure*}

Beyond the model's final answer, the latent tokens might be altering other parts of the final answer distribution. We analyze how the models’ answer distribution changes with more latent tokens by tracking Rank-Biased Overlap (p=0.90), Top-k Jaccard similarity (k=10), and Jensen-Shannon Divergence over the logit vectors. The results in \autoref{fig:distributional_similarity} tell a similar story as our other results: the model’s answer distribution (not just the top-1 prediction) is relatively stable with additional latent tokens for PrOntoQA and ProsQA. But on GSM8k-Aug, the answer distribution changes significantly as the number of latent tokens varies.

\FloatBarrier
\newpage
\section{Gold reasoning trace backtracking experiment}\label{sec:appendix-backtracking}

\subsection{Backtracking search pseudocode}\label{ssec:appendix-backtracking-pseudocode}

\begin{algorithm}[ht]
\caption{Backtracking Search for Reasoning Trace in Vocabulary Projections}
\label{alg:backtracking}
\KwIn{$T_{\mathrm{primary}}$: primary reasoning trace\\
\phantom{\textbf{Input:} }$T_{\mathrm{alt}}$: alternative valid reasoning traces\\
\phantom{\textbf{Input:} }$V$: top-$k$ vocabulary projections at latent token and answer positions}
\KwOut{Best matching tree, or $\emptyset$ if none found}
\BlankLine

$\textit{best} \leftarrow \{\}$ \tcp*{map from trace $\to$ best tree found}

\ForEach{trace $T \in \{T_{\mathrm{primary}}\} \cup T_{\mathrm{alt}}$}{
    $G \leftarrow \textsc{BuildDAG}(T)$\;
    \tcp*[f]{Edges: operand $\to$ result; merge nodes if result reappears as operand in later step}
    \BlankLine

    \If{$\mathrm{final\_answer} \notin \mathrm{top\text{-}}k$ of $V[\mathrm{answer\_position}]$}{
        \Return $\emptyset$\;
    }
    \BlankLine

    $\textit{partial\_trees} \leftarrow \{(\{\},\; \{\text{operands of final step}\})\}$ \tcp*{set of (assignment, available) pairs}
    $\textit{found\_trees} \leftarrow \emptyset$\;
    \BlankLine

    \For{$\mathrm{pos} \leftarrow \mathrm{last\_latent}$ \KwTo $\mathrm{first\_latent}$}{
        $\textit{new\_partial} \leftarrow \emptyset$\;
        \ForEach{$(\textit{assignment},\, \textit{available}) \in \textit{partial\_trees}$}{
            $\textit{matches} \leftarrow \textit{available} \,\cap\, \mathrm{top\text{-}}k(V[\mathrm{pos}])$\;
            \If{$\textit{matches} = \emptyset$}{
                $\textit{new\_partial} \leftarrow \textit{new\_partial} \,\cup\, \{(\textit{assignment},\, \textit{available})\}$ \tcp*{unchanged}
            }
            \ForEach{node $n \in \textit{matches}$}{
                $\textit{new\_assign} \leftarrow \textit{assignment} \,\cup\, \{n \to \mathrm{pos}\}$\;
                $\textit{new\_avail} \leftarrow \textit{available} \,\cup\, \{\text{operands of } n\} \setminus \{n\}$\;
                $\textit{new\_partial} \leftarrow \textit{new\_partial} \,\cup\, \{(\textit{new\_assign},\, \textit{new\_avail})\}$\;
            }
        }
        $\textit{partial\_trees} \leftarrow \textit{new\_partial}$\;
        \BlankLine
        \ForEach{$(\textit{assignment},\, \textit{available}) \in \textit{partial\_trees}$}{
            \If{all leaves of $G$ are in \textit{assignment}}{
                $\textit{found\_trees} \leftarrow \textit{found\_trees} \,\cup\, \{\textit{assignment}\}$\;
            }
        }
    }
    \BlankLine

    \If{$\textit{found\_trees} \neq \emptyset$}{
        $\textit{best}[T] \leftarrow$ tree with highest projection ranks and earliest positions\;
    }
}
\BlankLine

\tcp{Select best across traces (prefer primary)}
\uIf{$T_{\mathrm{primary}} \in \textit{best}$}{
    \Return $\textit{best}[T_{\mathrm{primary}}]$\;
}
\uElseIf{$\exists\, T \in T_{\mathrm{alt}}$ s.t.\ $T \in \textit{best}$}{
    \Return tree with highest projection ranks and earliest positions among $\textit{best}[T_{\mathrm{alt}}]$\;
}
\Else{
    \Return $\emptyset$\;
}
\end{algorithm}

\FloatBarrier
\newpage
\subsection{Backtracking experiment examples}\label{ssec:appendix-backtracking-examples}

\begin{figure*}[htbp]
    \centering
    \includegraphics[width=\textwidth]{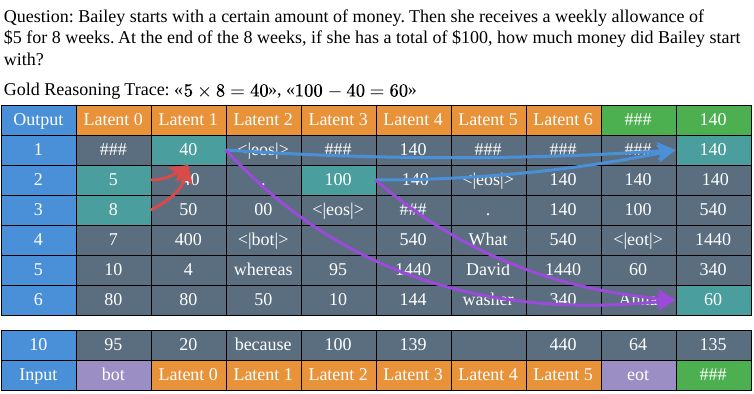}
    \caption{Found gold reasoning trace in Coconut + GPT-2 Small's vocabulary projections, from instance 69 of GSM8k-Aug's test split. The Coconut model encodes the correct final step, but it encodes an incorrect final step more strongly. The model seems to think that Bailey was losing \$5 per week, rather than receiving \$5 per week.}
    \label{fig:vp_table_sample_069_coconut_incorrect_GT_found}
\end{figure*}

\FloatBarrier
\subsection{Backtracking experiment error analysis}
When the backtracking search in \cref{sec:backtracking} fails to find an encoded gold reasoning trace, how is the Coconut model solving the problem? We find evidence against the worst case scenario, where the LRM arrives at the correct answer in a completely uninterpretable way. Instead, we find three main reasons why the backtracking search can fail even when the model gets the correct answer: a valid reasoning trace may be missing from the set of known reasoning traces, vocabulary projection does not encode multi-token concepts an easily identifiable way, or most of but not the entire gold reasoning trace may be encoded.

\begin{figure*}[htbp]
    \centering
    \includegraphics[width=\textwidth]{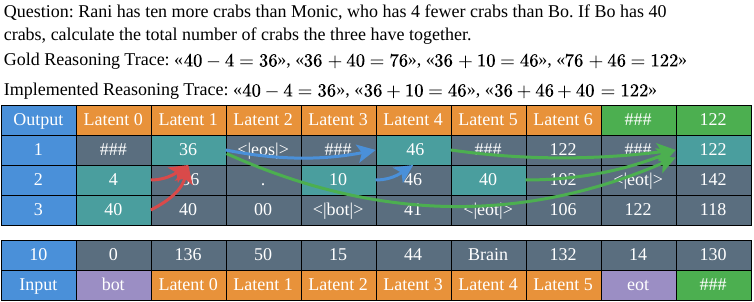}
    \caption{Coconut + GPT-2 Small's vocabulary projections, from instance 179 of GSM8k-Aug's test split. The Coconut model encodes a valid reasoning trace not contained in the set of known gold reasoning traces.}
    \label{fig:vp_table_sample_179_coconut_error_analysis}
\end{figure*}

\autoref{fig:vp_table_sample_179_coconut_error_analysis} shows an example where the model is following a valid reasoning trace that is not in the set of known reasoning traces. Specifically, the model skips Step 2 in the gold reasoning trace, which calculates $36+40=76$. Instead of calculating and storing this intermediate result, the Coconut model changes the last step from $76+46=122$ to an equivalent $36+40+46=122$. The MultiChain GSM8k-Aug dataset did not contain this alternative reasoning trace because its augmentations preserve the number of steps used in the original reasoning trace, and this modification reduces the step count from 4 to 3. To avoid this error, we'd need to enumerate all valid ways of solving the problem, which is impractical and often impossible.

\begin{figure*}[htbp]
    \centering
    \includegraphics[width=\textwidth]{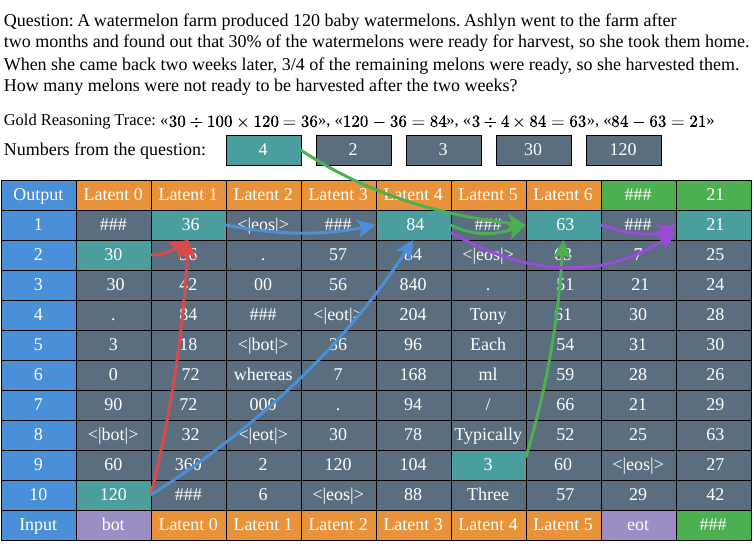}
    \caption{Coconut + GPT-2 Small's vocabulary projections, from instance 229 of GSM8k-Aug's test split. The model seems to encode the percentage 30\% as simply 30, instead of 30/100 as in the gold reasoning trace, causing the backtracking search to fail. The arrows indicate what the reasoning trace looks like when assuming the 30 is representing 30\%.}
    \label{fig:vp_table_sample_229_coconut_error_analysis}
\end{figure*}

\autoref{fig:vp_table_sample_229_coconut_error_analysis} shows an example where vocabulary projection limits our ability to identify decimals, percentages, and multi-token numbers generally that are encoded in a latent thought. The first step of the gold reasoning trace calculates 30\% of 120. The gold reasoning trace represents the 30\% as 30/100, which is equivalent, but the model does not represent the 100 in its vocabulary projection. Instead, it's likely that the ``30'' token in the top-2 of the vocabulary projection of the first latent token represents this percentage. But since the model and the gold reasoning trace represent this percentage differently, the backtracking search fails. 

\begin{figure*}[htbp]
    \centering
    \includegraphics[width=\textwidth]{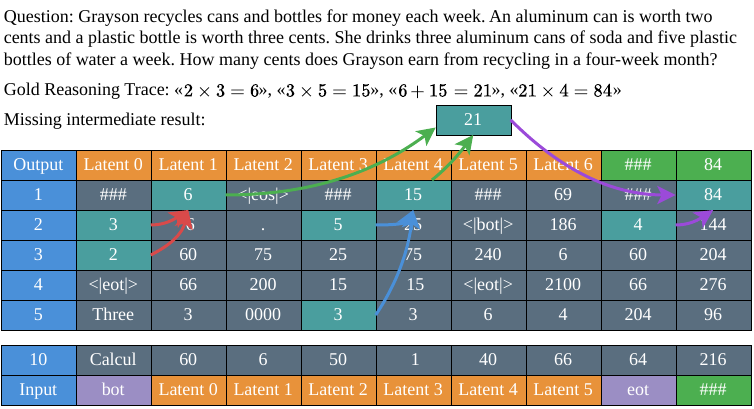}
    \caption{Coconut + GPT-2 Small's vocabulary projections, from instance 460 of GSM8k-Aug's test split. The gold reasoning trace is encoded, except for the intermediate result 21, which causes the backtracking search to fail.}
    \label{fig:vp_table_sample_460_coconut_error_analysis}
\end{figure*}

\autoref{fig:vp_table_sample_460_coconut_error_analysis} shows an example where Coconut's vocabulary projections contain a partial gold reasoning trace. In this instance, Step 3, $6+15=21$, is missing, so the intermediate result 21 is not encoded. However, the model is still able to calculate the result of the next and final step, 84. There are at at least two possibilities for this. It may encode the final step as $(6+15)*4=84$, which would make it more like a previously unknown valid reasoning trace. Or, it may be encoding 21 at the final reasoning position, and vocabulary projection incorrectly extracts it as 2100 and 210, which are shown in the table.

This error analysis suggests that our backtracking search results provide a lower bound on LRM interpretability, with failures stemming from methodological limitations rather than fundamentally uninterpretable model behavior. More robust reasoning trace finding techniques that handle equivalent reformulations and flexible numerical encodings would likely recover a higher proportion of encoded gold reasoning traces.

\FloatBarrier
\subsection{Backtracking experiment results by solution length}

\begin{figure}[htbp]
    \centering
    \includegraphics[width=1.0\textwidth]{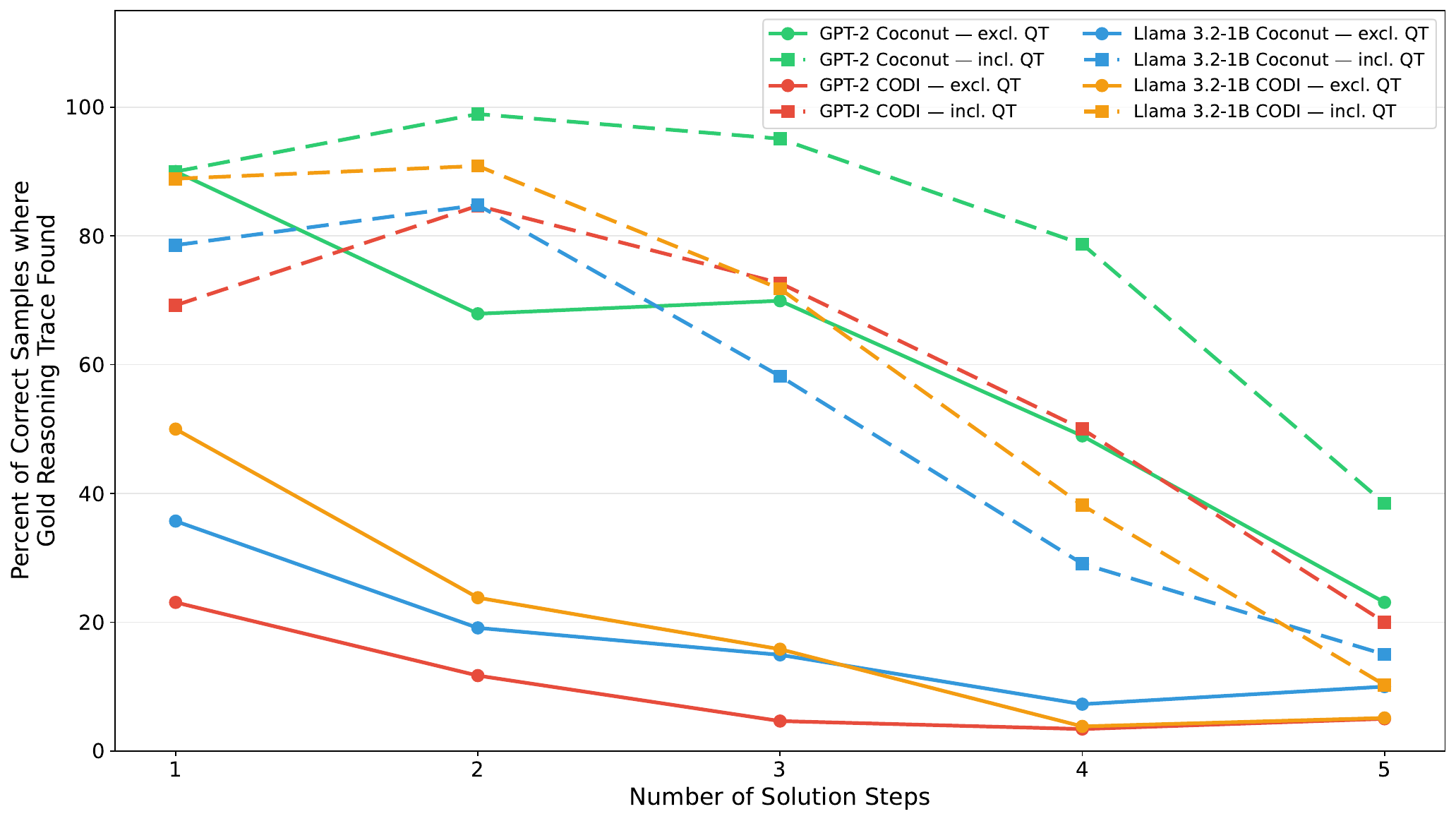}
    \caption{Percent of any gold reasoning trace found in the vocabulary projections of latent tokens for correctly answered problems by reasoning trace length. Reasoning traces with more than 5 steps are not shown due to low instance counts ($<3$).}
    \label{fig:back_tracking_vp_per_step_any_gold_label_multi_llm_combined}
\end{figure}

The representation of gold reasoning traces tends to decrease as the reasoning trace length increases beyond 3 steps, as shown in \autoref{fig:back_tracking_vp_per_step_any_gold_label_multi_llm_combined}. Including question tokens as potential operands, Coconut + GPT-2 Small declines from 99\% at two steps to 38\% at five steps. CODI + GPT-2 Small declines from 85\% at two steps to just 20\% at five steps. This degradation reflects a limitation in the models' capacity to maintain longer reasoning traces.

\FloatBarrier
\subsection{Incorrect predictions}

\begin{table}[h]
\begin{center}
\resizebox{\textwidth}{!}{
\begin{tabular}{lcccc}
\toprule
\multicolumn{1}{c}{\bf Method} &\multicolumn{1}{c}{\bf Base Model} & \multicolumn{1}{c}{\bf Incorrect Instances} & \multicolumn{1}{c}{\bf Correct Answer in Top-10} & \multicolumn{1}{c}{\bf Percent} \\
\midrule
Coconut & GPT-2 Small & 793 & 369 & 46.5\% \\
CODI    & GPT-2 Small & 675 & 337 & 49.9\% \\
Coconut & Llama-3.2-1B-Instruct & 759 & 425 & 56.0\% \\
CODI    & Llama-3.2-1B-Instruct & 511 & 275 & 53.8\% \\
\bottomrule
\end{tabular}
}
\end{center}
\caption{Incorrect predictions on GSM8k-Aug where the correct answer appears in the top-10 predicted tokens.}
\label{tab:incorrect-in-topk-all}
\end{table}

\FloatBarrier
\newpage
\section{Forward chaining experiment}\label{sec:appendix-forward-chaining}

\subsection{Forward chaining pseudocode}\label{ssec:appendix-forward-chaining-pseudocode}

\begin{algorithm}[ht]
\caption{Forward Chaining}
\label{alg:forward-chaining}
\KwIn{$V$: top-$k$ vocab projections at each latent token position and the answer positions\\
\phantom{\textbf{Input:} }$Q$: numbers extracted from the question\\
\phantom{\textbf{Input:} }$\mathrm{final\_answer}$: model's predicted answer\\
\phantom{\textbf{Input:} }$d$: position offset ($d{=}1$ for Coconut, $d{=}2$ for CODI)}
\KwOut{Computation tree and verification status}
\BlankLine
 
\For(\tcp*[f]{Phase 1: Generate candidate steps}){$\mathrm{pos} \leftarrow 0$ \KwTo $\mathrm{num\_latent\_positions}$}{
    $\textit{result} \leftarrow \text{top-1 integer at } V[\mathrm{pos}]$\;
    \lIf{$\textit{result}$ is None}{\textbf{continue}}
 
    $\textit{operands} \leftarrow \mathrm{top\text{-}}k \text{ integers at } V[\mathrm{pos} - d] \;\cup\; \{\text{top-1 integer at position } p : p < \mathrm{pos}\} \;\cup\; Q$\;
 
    $S_{2} \leftarrow \{(a, b, \mathrm{op}, \textit{result}) : a,b \in \textit{operands},\; a \;\mathrm{op}\; b = \textit{result},\; \mathrm{op} \in \{+,-,\times,\div\}\}$\;
    $S_{3} \leftarrow \{(a, b, c, \mathrm{op}_1, \mathrm{op}_2, \textit{result}) : a,b,c \in \textit{operands},\; a \;\mathrm{op}_1\; b \;\mathrm{op}_2\; c = \textit{result}\}$\;
    $\textit{steps} \leftarrow S_{2} \cup S_{3}$\;
 
    \tcp{Prioritize by: (1) operand source: verified intermediate $>$ question number $>$ top-$k$ $>$ unverified intermediate, (2) fewer operands}
    $\textit{candidates} \leftarrow \mathrm{sort}(\textit{steps}, \text{by priority above})$\;
 
    $\textit{best} \leftarrow \mathrm{None}$\;
    \ForEach(\tcp*[f]{Phase 2: Try to verify one candidate step}){$\textit{candidate} \in \textit{candidates}$}{
        \If{$\textsc{Verify}(\textit{candidate},\, n_{\mathrm{attempts}},\, r_{\mathrm{passes}})$}{
            $\textit{best} \leftarrow \textit{candidate}$\;
            \textbf{break}\;
        }
    }
    \lIf{$\textit{best} = \mathrm{None}$ \textbf{and} $\textit{candidates} \neq \emptyset$}{$\textit{best} \leftarrow \textit{candidates}[0]$}
}
\BlankLine
 
$\textit{root} \leftarrow \text{earliest step where } \textit{result} = \mathrm{final\_answer}$ \tcp*{Phase 3: Build reasoning trace}
$\textit{tree\_steps} \leftarrow \{\textit{root}\}$\;
\ForEach{step $\in$ \textit{tree\_steps}}{
    \ForEach{operand $\in$ step}{
        \If{operand came from a previous step's result}{
            $\textit{tree\_steps} \leftarrow \textit{tree\_steps} \,\cup\, \{\textit{source\_step}\}$\;
        }
    }
}
$\textit{tree\_verified} \leftarrow \mathrm{all}(\textit{step.verified} \text{ for } \textit{step} \in \textit{tree\_steps})$\;
\Return $\textit{tree\_steps}$ sorted by position, $\textit{tree\_verified}$\;
 
\hrulefill
\BlankLine
 
\SetKwProg{Fn}{Function}{:}{}
\Fn{\textsc{Verify}$(\textit{step},\, n_{\mathrm{attempts}},\, r_{\mathrm{passes}})$}{
    $\textit{pass\_count} \leftarrow 0$\;
    \For{$i \leftarrow 1$ \KwTo $n_{\mathrm{attempts}}$}{
        $\textit{var} \leftarrow \text{select operand traceable to question}$\;
        $\textit{new\_val} \leftarrow \text{sample different single-token integer}$\;
        $\textit{expected} \leftarrow \text{recompute step result with } \textit{new\_val}$\;
        $\textit{observed} \leftarrow \text{top-1 integer at } V'[\textit{step.position}]$ \tcp*{$V'$ from modified prompt}
        \lIf{$\textit{observed} = \textit{expected}$}{$\textit{pass\_count} \leftarrow \textit{pass\_count} + 1$}
    }
    \Return $\textit{pass\_count} \geq r_{\mathrm{passes}}$\;
}
\end{algorithm}

\FloatBarrier
\newpage
\subsection{Forward chaining verification example}\label{ssec:appendix-verification-example}

\begin{figure*}[htbp]
    \centering
    \includegraphics[width=\textwidth]{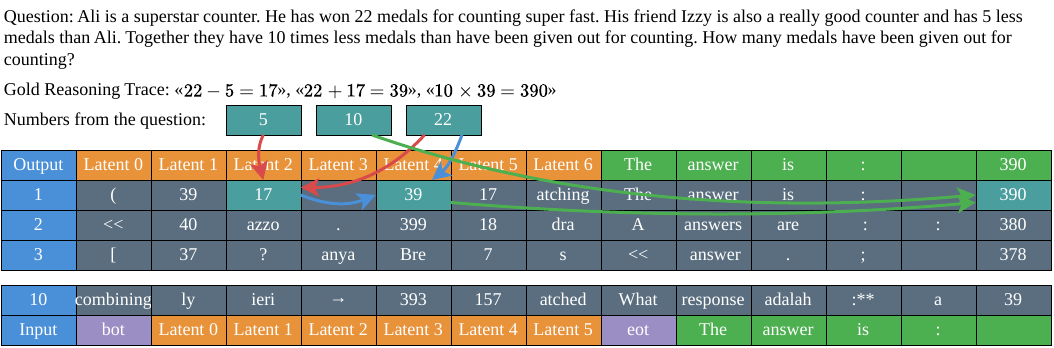}
    \caption{CODI + Llama-3.2-1B-Instruct's vocabulary projections, from instance 290 of GSM8k-Aug's test split. The reasoning trace found and verified by the forward chaining method is displayed. This reasoning trace happens to match the gold reasoning trace.}
    \label{fig:vp_table_sample_290_codi_llama}
\end{figure*}

This section contains an example of how the forward chaining method works. \autoref{fig:vp_table_sample_290_codi_llama} shows a found and verified reasoning trace, which happens to be the same as the gold reasoning trace for this instance. First, we generate candidate steps that may be encoded for each latent token. Latent token 0 has no candidate steps: it has no integer tokens in its top-10 vocabulary projection, so there is no step result. Latent token 1 also has no candidate steps. Its top integer token is 39, but there's no arithmetic combination of candidate operands that can combine to produce it. 

\begin{figure*}[htbp]
    \centering
    \includegraphics[width=\textwidth]{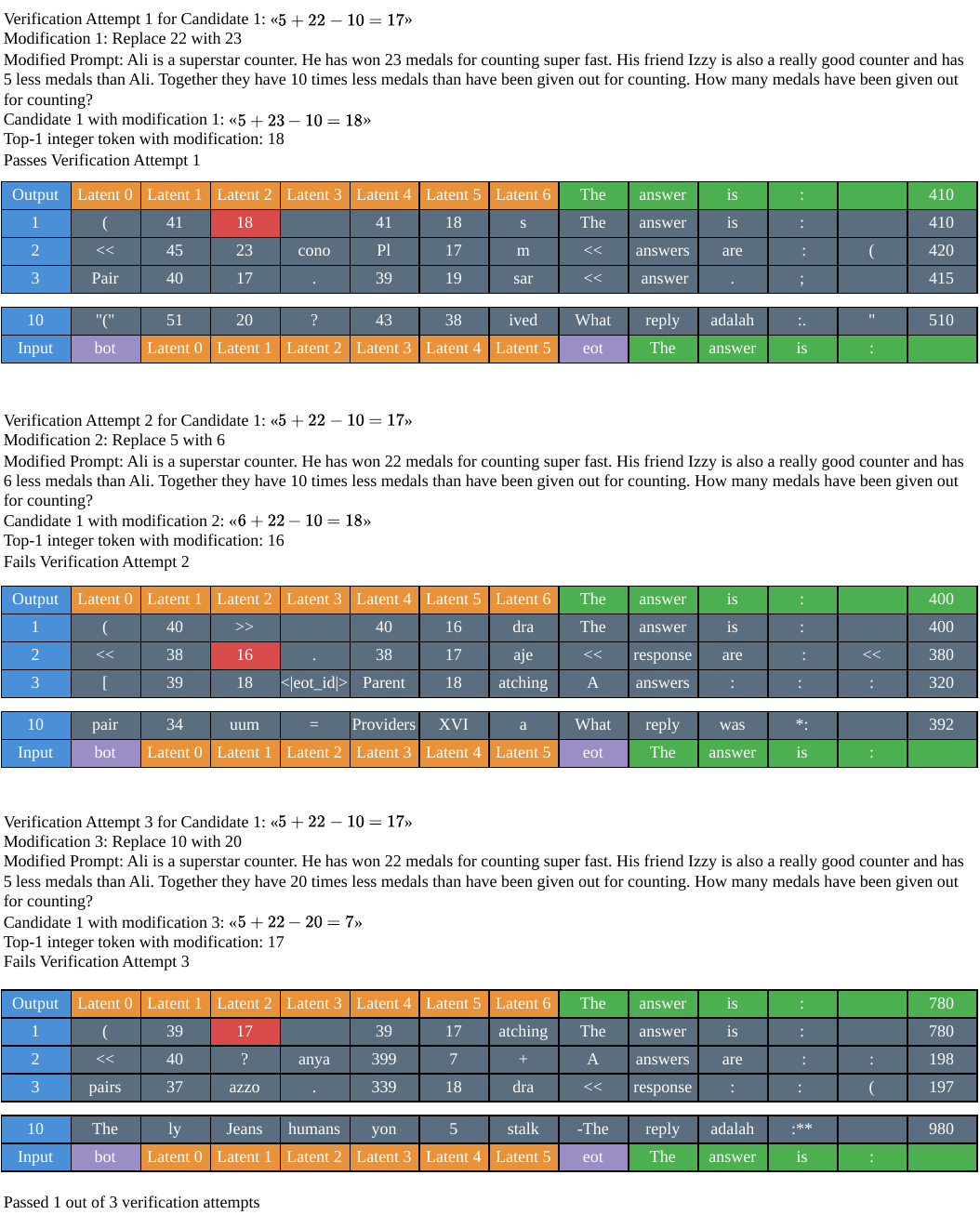}
    \caption{Verification process for latent token 2, candidate step 1 from instance 290 of GSM8k-Aug's test split.}
    \label{fig:sample_290_codi_llama_LT2_candidate1}
\end{figure*}

\begin{figure*}[htbp]
    \centering
    \includegraphics[width=\textwidth]{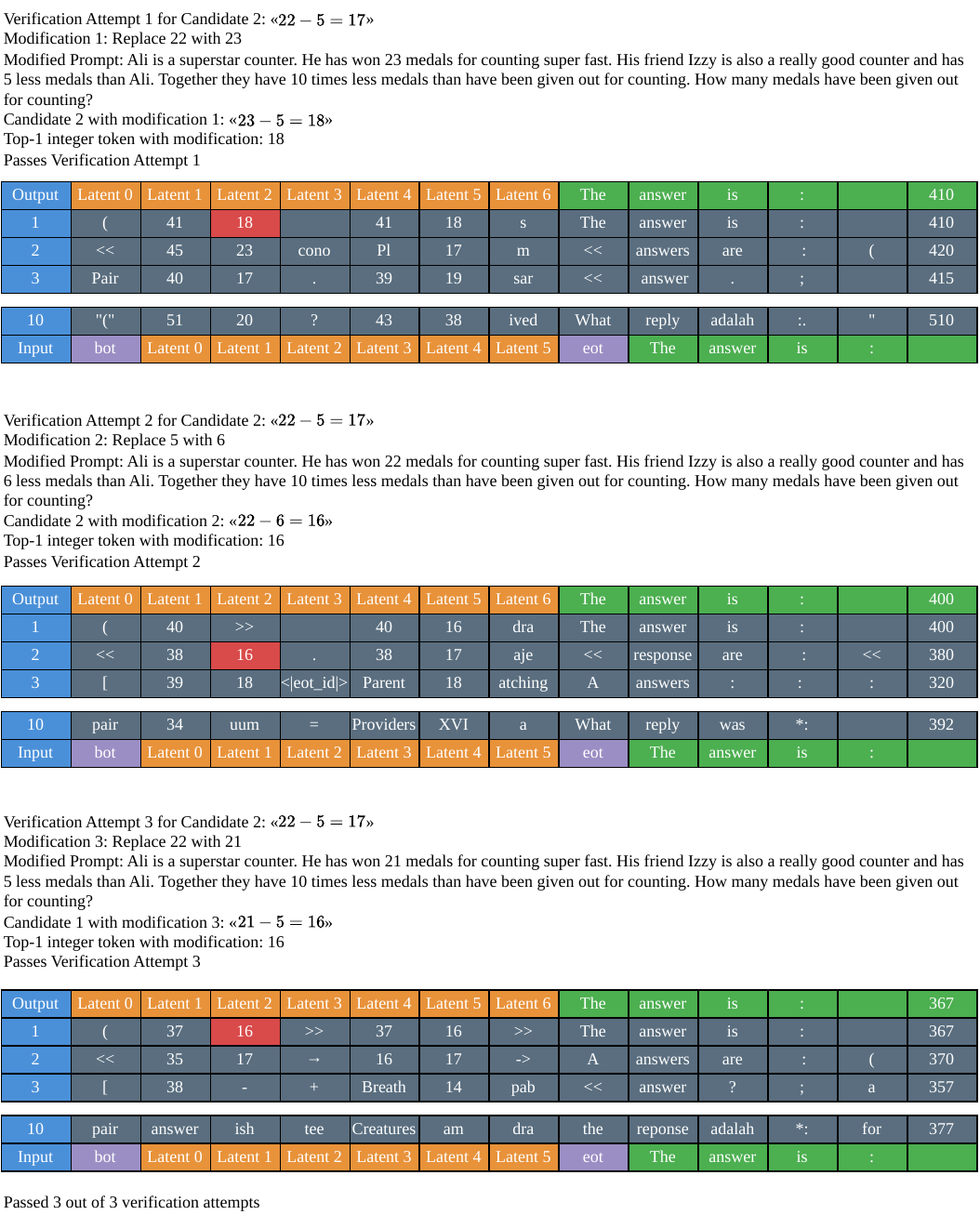}
    \caption{Verification process for latent token 2, candidate step 2 from instance 290 of GSM8k-Aug's test split.}
    \label{fig:sample_290_codi_llama_LT2_candidate2}
\end{figure*}

Latent token 2's top integer token is 17, so we assume that is the result produced by the encoded step. There are two candidate steps that the model may be using to get 17: $5 + 22 - 10 = 17$ and $22 - 5 = 17$. The first candidate step passes 1 out of 3 verification attempts, as shown in \autoref{fig:sample_290_codi_llama_LT2_candidate1}. The second candidate step passes 3 out of 3 verification attempts, as shown in \autoref{fig:sample_290_codi_llama_LT2_candidate2}. 

The process continues for the remaining latent tokens and the answer position. It verifies the step $22 + 17 = 39$ at latent token 4 with 3 out of 3 verification attempts passing, and $39 * 10 = 390$ at the answer position with 2 out of 3 verification attempts passing. The forward chaining method then assembles the found steps into the full reasoning trace: $22 - 5 = 17$, $22 + 17 = 39$, and $10 * 39 = 390$. This reasoning trace is considered verified for 1 or 2 required passes, and unverified for 3 required passes.

\FloatBarrier
\newpage
\subsection{Dataset requirements for the forward chaining method}\label{ssec:appendix-forward-chaining-requirements}

The forward chaining method described in \cref{sec:forward_chaining} will work with a dataset that meets the following requirements:

\begin{enumerate}
    \item The reasoning trace must be decomposable into steps. 
    \item Each step must be a deterministic function of its operands and produce one result.
    \item The operators must be a known, small set so that forward chaining can brute-force search over them.
    \item The operands and results must be single-token, so that they can be observed using vocabulary projection. Note that this is also a function of the tokenizer used. 
    \item For each step, at least one base operand must be mentioned in the prompt. The base operands are just the step operands, or, if one of the operands is the result of a previous step, then the base operands can be the base operands of that previous step. A reasoning step cannot be fully based on operands from its world knowledge. If no base operands are mentioned in the prompt, then this method cannot verify the step by modifying the prompt.
    \item The base operands and step results must be distinguishable from each other. This makes it unambiguous which base operand mentioned in the prompt should be modified to verify a given step.
\end{enumerate}

Requirement 3 can be removed if future work finds a way to detect the operator used from the model's representations directly. In our experiments, we found that the LRMs studied do not encode the operators in the latent tokens, at least not in a way that is detectable using vocabulary projection. Since there are only four operators used in GSM8k-Aug, we brute-force search over them.

Datasets can be modified to meet requirement 4 by replacing key multi-token concepts with single-token concepts. Alternatively, future work could use a method other than vocabulary projection for detecting concepts encoded in a latent token, which would allow for multi-token operands and intermediate results.

Requirement 6 makes this method difficult to use for tasks that have a small set of possible operands or results, which make it likely for multiple steps to have the same intermediate result. E.g., the step results in logical reasoning tasks may be only ``True'' or ``False.''

\FloatBarrier
\newpage
\section{PrOntoQA heuristic}\label{sec:appendix-prontoqa}

\begin{figure*}[htbp]
    \centering
    \includegraphics[width=\textwidth]{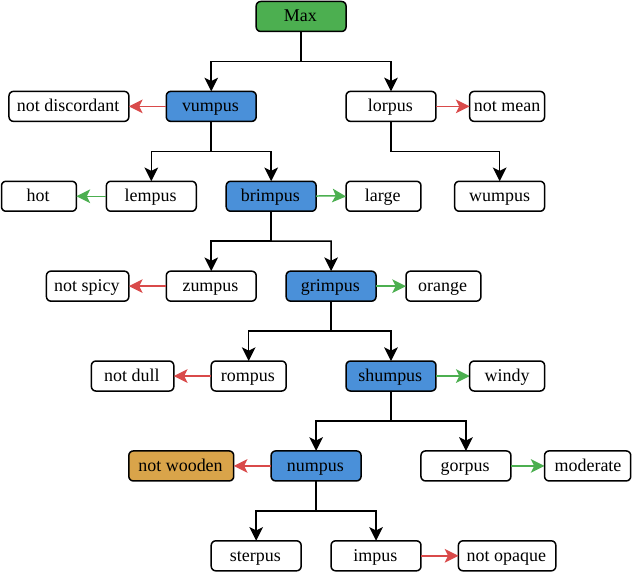}
    \caption{Instance 1 of the train split of PrOntoQA shown as a directed acyclic graph. This instance is also shown in text in \autoref{fig:dataset-examples}.}
    \label{fig:prontoqa_train_1}
\end{figure*}

During our investigation into how models solve PrOntoQA, we noticed that all instances from this dataset can be solved by treating the problem as a directed acyclic graph and continually exploring the child node with the most children nodes until you arrive at the node with the property in question. This can be implemented as a counting task for the child node that is mentioned the most in the prompt. E.g., for the problem in \autoref{fig:prontoqa_train_1}, starting at the start node named ``Max'', there are two choices of child nodes: vumpus and lorpus. Vumpus has two children, and lorpus has one. Since vumpus has more children, it is also mentioned more frequently in the promt than lorpus. You can keep going down the graph in the same manner until numpus is reached, which has the property in question of ``not wooden''. Since all instances of PrOntoQA can be solved in this way, it's possible for models to exploit this heuristic instead of learning to search or do first-order logical reasoning. 

\end{document}

%% file: math_commands.tex
\usepackage{amsmath,amsfonts,bm}

\def\eqref#1{equation~\ref{#1}}

\def\1{\bm{1}}

\DeclareMathAlphabet{\mathsfit}{\encodingdefault}{\sfdefault}{m}{sl}
\SetMathAlphabet{\mathsfit}{bold}{\encodingdefault}{\sfdefault}{bx}{n}

